%% file: main.tex
\documentclass{article} % For LaTeX2e
\usepackage{iclr2024_conference,times}

\input{math_commands.tex}

\usepackage{makecell}
\usepackage[pagebackref=true,breaklinks=true,citecolor=blue,linkcolor=blue,colorlinks,urlcolor=blue,bookmarks=false]{hyperref}
\usepackage{url}
\usepackage{booktabs}
\usepackage{arydshln}
\usepackage{multirow}
\usepackage{graphicx}
\usepackage{subcaption}
\usepackage{enumitem}
\usepackage{xspace}
\usepackage{mathabx}
\usepackage{bbm}
\usepackage{wrapfig}
\usepackage{xcolor}

\graphicspath{{figures/}}

\usepackage[capitalize]{cleveref}
\crefname{section}{Section}{Secs.}
\Crefname{section}{Section}{Sections}
\Crefname{table}{Table}{Tables}
\crefname{table}{Tab.}{Tabs.}

\makeatletter
\DeclareRobustCommand\onedot{\futurelet\@let@token\@onedot}
\def\@onedot{\ifx\@let@token.\else.\null\fi\xspace}

\def\eg{\emph{e.g}\onedot} 
\def\ie{\emph{i.e}\onedot} 
\def\cf{\emph{c.f}\onedot} 
\def\etc{\emph{etc}\onedot} \def\vs{\emph{vs}\onedot}

\makeatother

\definecolor{mygray}{gray}{0.4}
\newenvironment{mytiny}{\color{mygray}\small}{}

\newcommand{\quantizername}{LFQ}
\newcommand{\modelname}{MAGVIT-v2}
\newcommand{\webpage}{{\small \url{https://magvit.cs.cmu.edu/v2}}}

\title{Language Model Beats Diffusion\\---  Tokenizer is Key to Visual Generation}

\author{%
\hspace{-3mm}
\small
\noindent
\textbf{Lijun Yu}$^{\ddagger \dagger}$\thanks{Work done during a research internship at Google Research.} \hfill 
\textbf{Jos\'e Lezama}$^{\dagger}$ \hfill 
\textbf{Nitesh B. Gundavarapu}$^{\dagger}$  \hfill 
\textbf{Luca Versari}$^{\dagger}$ \hfill 
\textbf{Kihyuk Sohn}$^{\dagger}$
\vspace{0.3em}
\\
\hspace{-3mm}
\small
\noindent
\textbf{David Minnen}$^{\dagger}$ \hfill 
\textbf{Yong Cheng}$^{\dagger}$ \hfill 
\textbf{Vighnesh Birodkar}$^{\dagger}$ \hfill 
\textbf{Agrim Gupta}$^{\dagger}$ \hfill 
\textbf{Xiuye Gu}$^{\dagger}$ \hfill
\textbf{Alexander G. Hauptmann}$^{\ddagger}$
\vspace{0.3em}
\\
\hspace{-3mm}
\small
\noindent
\textbf{Boqing Gong}$^{\dagger}$ \hfill 
\textbf{Ming-Hsuan Yang}$^{\dagger}$ \hfill 
\textbf{Irfan Essa}$^{\dagger}$ \hfill
\textbf{David A. Ross}$^{\dagger}$ \hfill 
\textbf{Lu Jiang}$^{\dagger \ddagger}$
\vspace{0.3em}
\\
$^\dagger$Google, $^\ddagger$Carnegie Mellon University 
\vspace{-5mm}
}

\newcommand{\cl}[1]{{#1}}

\iclrfinalcopy % Uncomment for camera-ready version, but NOT for submission.
\begin{document}

\maketitle
\vspace{-3mm}
\begin{abstract}
\vspace{-3mm}
While Large Language Models (LLMs) are the dominant models for generative tasks in language, they do not perform as well as diffusion models on image and video generation.
% their performance for image and video generation is not as good as falls behind that of diffusion models.
To effectively use LLMs for visual generation, one crucial component is the visual tokenizer that maps pixel-space inputs to discrete tokens appropriate for LLM learning.
% One crucial component for using LLMs effectively in visual generation is obtaining a good visual tokenizer, mapping pixel-space inputs to discrete visual tokens, such that the resulting tokens are appropriate for LLM modeling. 
In this paper, we introduce \modelname{}, a video tokenizer designed to generate concise and expressive tokens for both videos and images using a common token vocabulary.
% Our tokenizer utilizes a novel quantization approach and supports tokenization of both images and videos using a shared vocabulary
%that eliminates the need for codebook lookup, generating binary codes directly. Furthermore, architectural modifications lead to not only quality enhancements but also tokenization of both images and videos using a shared vocabulary.
Equipped with this new tokenizer, we show that LLMs outperform diffusion models on standard image and video generation benchmarks including ImageNet and Kinetics. 
In addition, we demonstrate that our tokenizer surpasses the previously top-performing video tokenizer on two more tasks: (1) video compression comparable to the next-generation video codec (VVC) according to human evaluations, and (2) learning effective representations for action recognition tasks.
% serving as an effective pre-training target for video action recognition.
%In addition to generation qiality, we show two potential brought by the proposed tokens: it produces favoarble compression than the
% The proposed tokenizer significantly outperforms previous video tokenizers and constitutes the first neural model to outperform the next-generation video codec H.266 in human evaluation. Furthermore, we demonstrate that the obtained representations are effective for self-supervised pre-training for video classification.
\end{abstract}

\vspace{-4mm}
\section{Introduction}
\vspace{-2mm}

\input{sections/introduction} \label{sec:intro}

\vspace{-4mm}
\section{Background} \label{sec:background}
\input{sections/preliminary}

\vspace{-4mm}
\section{Method}
\vspace{-2mm}
\input{sections/method}
\vspace{-4mm}
\section{Experiments}
\vspace{-2mm}
\input{sections/experiments}

\vspace{-4mm}
\section{Related Work}
\input{sections/related}

\vspace{-4mm}
\section{Conclusion and Future Work}
\vspace{-3mm}
\input{sections/conclusion}

\subsubsection*{Acknowledgments}
We would like to express our gratitude to Yu-Chuan Su and Sergey Ioffe for their valuable comments on our work, to Josh Dillon for discussions, 
% to Vighnesh Birodkar for help in baselineimplementation,
and to Eirikur Agustsson for help in compression evaluation.

\bibliography{reference}
\bibliographystyle{iclr2024_conference}

\clearpage

\appendix
%\section{Appendix}
\input{sections/appendix}

\end{document}

%% file: math_commands.tex
\usepackage{amsmath,amsfonts,bm}

\def\eqref#1{equation~\ref{#1}}
\def\1{\mathbbm{1}}

\def\rvc{{\mathbf{c}}}

\def\rvm{{\mathbf{m}}}

\def\rvx{{\mathbf{x}}}

\def\rvz{{\mathbf{z}}}

\def\ervm{{\textnormal{m}}}

\def\ervx{{\textnormal{x}}}

\def\ervz{{\textnormal{z}}}

\def\rmC{{\mathbf{C}}}

\def\rmV{{\mathbf{V}}}

\def\rmX{{\mathbf{X}}}

\def\rmZ{{\mathbf{Z}}}

\def\vone{{\bm{1}}}

\def\vs{{\bm{s}}}

\DeclareMathAlphabet{\mathsfit}{\encodingdefault}{\sfdefault}{m}{sl}
\SetMathAlphabet{\mathsfit}{bold}{\encodingdefault}{\sfdefault}{bx}{n}

\def\gL{{\mathcal{L}}}

\def\sC{{\mathbb{C}}}

\def\sR{{\mathbb{R}}}

\newcommand{\E}{\mathbb{E}}

\DeclareMathOperator{\argmin}{arg\,min}

\DeclareMathOperator{\sign}{sign}

%% file: sections/introduction.tex
% Generative AI is popular.
% There exist several different models for content generation: 

%The language model, commonly referred to as LM or LLM, is the de facto model for natural language modeling
Large transformer-based language models, commonly referred to as LMs or LLMs, are the de facto models for natural language generation~\citep{openai2023gpt4,googlepalm2,touvron2023llama}. Over time, LMs have expanded their capabilities to generate content in various modalities, asserting their dominance in other domains like audio~\citep{agostinelli2023musiclm}, speech~\citep{rubenstein2023audiopalm}, code generation~\citep{li2023starcoder}, medical applications~\citep{singhal2023towards} and robotics~\citep{brohan2023rt}.

% LMs have established its leading role in various domains beyond natural language such as .
%\lu{add more audio generation paper}.
% The modality (\eg, audio or image) is mapped into discrete tokens, which are treated as language words within the transformer architecture.

LMs are capable of generating images and videos. To do so, the image pixels are mapped into a sequence of discrete tokens by a visual tokenizer (\cf \cref{sec:background}). These tokens are then fed into the LM transformer, as if they were lexical words, for generative modeling. Despite notable advancements in employing LMs for visual generation~\citep{esser2021taming,chang2022maskgit}, LMs still do not perform as well as diffusion models~\citep{rombach2022high}.
% operate on sequences of visual tokens obtained by some form of vector-quantized autoencoder (\cref{sec:background})
% In the case of visual (image or video) generation, the best performing language models operate on sequences of visual tokens obtained by some form of vector-quantized autoencoder (\cref{sec:background}). Although there has been significant progress in using LMs under this setting \citep{esser2021taming,chang2022maskgit,yu2022magvit}, LMs still trail behind diffusion models~\citep{rombach2022high}. 
For instance, when evaluating on the ImageNet dataset, a gold standard benchmark for image generation, the best language model~\citep{lee2022draft} underperforms 
% in comparison to
the diffusion model~\citep{gao2023masked} by a substantial 48\% margin (FID 3.41 \vs 1.79 when generating images at the 256$\times$256 resolution).

\emph{Why do language models lag behind diffusion models in visual generation?} This paper suggests that a primary reason is the lack of a good visual representation, resembling our natural language system, for effectively modeling the visual world.
% Unlike natural language, humans have not yet developed an optimal vocabulary for our visual world. The challenges of learning such visual vocabulary significantly limit the generative capabilities of the LLM, which otherwise excel in language, graph, and various other domains.
To substantiate this hypothesis, this paper shows that, when utilizing a good visual tokenizer, the masked language model~\citep{devlin2019bert,chang2022maskgit,yu2022magvit} surpasses the state-of-the-art diffusion models  in terms of both generation fidelity and efficiency across image and video benchmarks, given the same training data, comparable model size, and training budget. To the best of our knowledge, this provides the first evidence that language models beat diffusion models on the hallmark ImageNet benchmark. 

% To the best of our knowledge, our results provide the first evidence suggesting that the language model can outperform the diffusion model on ImageNet given the same training data, comparable model size, and training budget. This progress is attributed to the use of a good visual tokenizer. 

It is worth emphasizing that our intention is not to assert whether the language model is superior to others, but to promote the exploration of visual tokenization methods for LLMs.
A fundamental difference of LLMs from other models, such as diffusion models, is that LLMs utilize a discrete latent format: tokens obtained from a visual tokenizer.
%MH: argue -> show (argue is a bit aggressive and confrontional)
We show that the values of these discrete visual tokens should not be overlooked considering their distinct advantages as follows. \textbf{(1) Compatibility with LLMs.} The main advantage of a token representation is that it shares the same form as language tokens, making it straightforward to leverage the optimizations our community has developed over many years for LLMs. This includes faster training and inference speeds~\citep{shazeer2019fast,lester2021power}, advancements in model infrastructure~\citep{dao2022flashattention,du2022glam}, learning recipes for model scaling~\citep{brown2020language,chowdhery2022palm}, and GPU/TPU optimization, among other innovations. Unifying vision and language by the same token space could set the stage for a true multimodal LLM that can understand, generate, and reason within our visual environment.
%as illustrated by AudioPaLM~\citep{rubenstein2023audiopalm} in the audio domain. 
% the biggest benefit of token representation resembeles of lanuage, rendering it straightforward to leverage all the optimization methods, the community has been deleopved for years, for the LLMs include training and inference speed~\citep{shazeer2019fast,ainslie2023gqa}, , even hardware GPU/TPU among others.
%\lu{yong} Could you please provide some references for well-known efforts to enhance the training and inference. Works have shown joint learning of understanding and generation suggest feasible~\citep{rubenstein2023audiopalm,yu2023scaling}.
%MQA and GQA~\citep{shazeer2019fast,ainslie2023gqa} introduce architectural modifications aimed at boosting the training and inference speed of the Transformer model. Some other initiatives focus on refining fine-tuning efficiency by incorporating slight parameter adjustments to a single model to cater to various scenarios~\citep{hu2021lora,lester2021power}.
% efficiency of large language models (LLMs), including optimizations for GPU/TPU hardware utilization in LLMs? [here we do not consider acceleration technique such as distillation]
\textbf{(2) Compressed representation.} The discrete token may offer a fresh perspective on video compression. The 
visual tokens can serve as a new video compression format to reduce disk storage and bandwidth during internet transfers. Unlike compressed RGB pixels, these tokens can be fed directly into generative models, bypassing the conventional decompression and latent encoding steps. This allows for faster processing in generative video applications, especially beneficial in edge computing cases. \textbf{(3) Visual understanding benefits}. 
Prior research has shown that the discrete tokens are valuable as a pre-training target in self-supervised representation learning, as discussed in BEiT~\citep{bao2021beit} and BEVT~\citep{wang2022bevt}. 
Additionally, research finds that using tokens as the model inputs improves the robustness and generalization~\citep{mao2021discrete}.

% However, the quality of transformer (language model and masked token model) is lagging behind denoising diffusion on image and video generation. For example, on the standard ImageNet benchmark, the best diffusion gets xx\% better FID=?? versus FID=6.8. On the other hand, LLM has been a de-facto model in the field of language modeling, and has established its leading role audio generation~\cite{}. What's missing? 

In this paper, we introduce \modelname{}, a video tokenizer designed to map videos (and images) into compact discrete tokens. Our model is built on the state-of-the-art video tokenizer, MAGVIT~\citep{yu2022magvit}, within the VQ-VAE framework~\citep{van2017neural}. We propose two new techniques. First, a novel lookup-free quantization method enables the learning of a large vocabulary that is able to improve generation quality of the language model.
Second, through extensive empirical analyses, we have identified modifications to the tokenizer that not only enhance generation quality but also enable the tokenization of both images and videos using a shared vocabulary.
%.
%enables the learning of a large vocabulary that is able to
% his new method eliminates conventional need for codebook lookup and produces binary codes directly. 
%By virtual of this design, the generation quality of the language model (LM) significantly enhances over a large vocabulary, a property not seen in existing VQ-VAE approaches. 

% Our primary focus is on video tokenization, which challenges in spatial-temporal compression.
% that can effectively map video (and image) into compact tokens. We focus on the video tokenization which is challenging to the spatial-temporal compression. 
%\lu{explains the new designs}; \lu{two novel designs tokenizer}. 

% enables the masked language model to yield higher fidelity compared to the state-of-the-art diffusion model on the ImageNet benchmark, using a similar model size and training budget. It also significantly outperforms the diffusion model as well as the prior best tokenizer on the standard video benchmark. 

We empirically demonstrate that our model outperforms the previously top-performing video tokenizer, MAGVIT, in three key areas. First, our model significantly improves the generation quality of MAGVIT, establishing the state of the art on the common image and video benchmarks. Second, user studies indicate that its compression quality exceeds that of MAGVIT and the current video compression standard, HEVC~\citep{sullivan2012overview}. Moreover, it is on par with the next-generation video codec, VVC~\citep{vvc}.
Finally, we show that, compared to MAGVIT, our new tokens are stronger for video understanding tasks across two setups and three datasets.
%Kinetics~\citep{carreira2018short} and Something-Something V2~\citep{goyal2017something}. 
The main contributions of this work are:
\vspace{-2mm}
\begin{itemize}[nosep, leftmargin=*]
\item A new video tokenizer that outperforms the previously best-performing video tokenizer in three areas: visual generation, video compression, and action recognition.
\item A novel lookup-free quantization approach that enables improving the visual generation quality of language models by learning a large vocabulary.
\item To the best of our knowledge, the first evidence suggesting that a language model can outperform diffusion models on ImageNet when provided with the same training data, an equivalent model size, and a similar training budget.
\item A video compressor with better quality than HEVC and VVC, at similar bit rates, according to user studies. To our knowledge, this is the first successful attempt of a visual tokenizer designed for video generation to achieve comparable results to standard codecs.
% Generation: we achieve state-of-the-art video and image generation with token-based transformers that can seamlessly utilize existing LLM infrastructure. This also marks the first time where token-based transformers perform favorably against diffusion models on the important ImageNet benchmarks.
% \item Compression: we accomplish high-fidelity video reconstruction better than prior best video tokenizer ~\citep{yu2022magvit}. In addition, we present the first neural model that outperforms next-generation video codec H.266 in human evaluation.
% \item Understanding: we demonstrate the effectiveness of using tokens for pretraining which benefits video classification performance.
\end{itemize}

%% file: sections/preliminary.tex
\vspace{-4mm}
\paragraph{Language Model (LM) for visual generation.}
LMs have been extended to generate images and videos. A visual tokenizer $f$ is used to first map visual inputs into a sequence of discrete tokens. A video  $\rmV \in \sR^{T \times H \times W \times 3}$ (or image when $T=1$) is tokenized into a discrete representation $\rmX = f(\rmV) \in \{1, 2, \cdots, K\}^{T' \times H' \times W'}$, where $K$ is the codebook (vocabulary) size of the visual tokenizer. $\rmX$ is flattened into a 1D token sequence obtained using raster scan ordering and then fed into an LM transformer for generative modeling. 

Two types of LMs are commonly used for visual generation. The \emph{Autoregressive LM (AR-LM)} includes ImageGPT~\citep{chen2020generative}, DALL-E~\citep{ramesh2021zero}, Parti~\citep{yu2022scaling}, \etc. 
An AR-LM predicts the next token given the previous tokens along with additional conditioning information $\rvc$ using a categorical distribution for $p_\theta(\ervx_i \mid \rvx_{<i}; \rvc)$. During inference, AR-LMs use the standard autoregressive decoding over the tokens. Finally, the tokens are converted back to pixels by a decoder associated with the visual tokenizer.

%A new sequence can be directly sampled from the model by $\hat{\rvx}_i \sim p_\theta(\rvx_i | \hat{\rvx}_0, \ldots, \hat{\rvx}_{i-1}, \rvc)$.

The \emph{Masked LM (MLM)} is another type of language model for visual generation, such as: MaskGIT~\citep{chang2022maskgit}, MAGVIT~\citep{yu2022magvit}, Phenaki~\citep{villegas2022phenaki}, and MUSE~\citep{chang2023muse}, among others. An MLM is trained using a masked token objective \citep{devlin2019bert}, where some tokens in the sequence are randomly masked and need to be predicted given the observed tokens. Let $\rvm \in \{0,1\}^n$ be a random binary sequence where $\rvm^\top \vone \in [0, n-1]$. The MLM learns $p_\theta(\ervx_i \mid \{\ervx_j: \ervm_j=1, \forall j\}; \rvc)$ for all $i$ where $\rvm_i=0$. To generate a video or image during inference, the MLM uses the non-autoregressive decoding algorithms
%\citep{ghazvininejad2019mask} with adaptations tailored 
for images and videos~\citep{chang2022maskgit,yu2022magvit}. 
The decoding starts with a fully masked sequence, which is iteratively filled by repeating two steps: (1) sample the whole sequence $\hat{\rvx}^{(t)}$ from $p_\theta$ given the non-masked tokens from the previous step, (2) re-mask the $\lfloor\lambda(t)\cdot n\rfloor$ tokens in $\hat{\rvx}^{(t)}$ with the lowest probability, following a decreasing masking ratio schedule $\lambda(t)$, according to timestamp $t$.
% For image algorithm details, refer to \citet{chang2022maskgit}; for video, see \citet{yu2022magvit}. 

\vspace{-4mm}
\paragraph{Denoising Diffusion Models (DDM).}
DDMs \citep{sohl2015deep,song2019generative} are regarded as the state-of-the-art in visual generation due to their high-quality image~\citep{dhariwal2021diffusion,ho2022imagen} and video generation~\citep{ho2022video}. For instance, DDPM~\citep{ho2020denoising} learns a denoising process parameterized as conditional Gaussian distributions over image pixels. Recently, diffusion models and language models have displayed a significant overlap. Recent DDMs diffuse over latents rather than raw pixels. These latents are obtained using models similar to the visual tokenizer used by LMs. In fact, the very first latent in diffusion, proposed by \citet{rombach2022high}, is derived from a visual tokenizer. Additionally, the diffusion model's architecture has been shifting from the U-Net to the transformer architecture~\citep{peebles2022scalable}. Consequently, the boundaries between diffusion and language models in visual generation have become less distinct. 
%Specifically, both DDM and MLM emulate a gradual corruption process on the latents and generate samples through iterative denoising restoration.
Yet, a fundamental difference between DDMs and LMs lies in the latent format, \ie, continuous \vs discrete. We have discussed the benefits of having discrete tokens in \cref{sec:intro} and 
%: compatibility with language models, a compact representation, and a video pretraining target \lijun{todo}. We 
will show that the proposed tokenizer improves in these aspects.
%Both models emulate a gradual corruption process, whether through random masking or random additive noise, which ends in a simple distribution (Dirac delta on all-mask sequence or standard Gaussian). In both cases, sampling is achieved via iterative restoration.

% a \emph{continuous} latent space \citep{rombach2022high}. We focus on the latter since it is more closely related to the tokenization approach. Connections between NAR models and DDMs for visual generation are very tight: both model a gradual corruption process (random masking or random additive noise) that ends in a simple distribution (Dirac delta on all-mask sequence or standard Gaussian), and sampling is done by iterative restoration. Furthermore, the original latent diffusion of \citep{rombach2022high} used the pre-quantization latent space of a VQ tokenizer. This boundary is made even more blurry by the convergence of both models to using a Transformer backbone (\citep{peebles2022scalable}).

\vspace{-4mm}
\paragraph{Visual tokenization.}
% \label{sec:background_tokenizer}
Visual tokenization plays an essential role in mapping pixels into a discrete representation suitable for generative modeling. VQ-VAE \citep{van2017neural} is a cornerstone work in image tokenization. A VQ-VAE model consists of a convolutional neural network (CNN) encoder, a vector-quantization (VQ) bottleneck, and a CNN decoder.
Given a video $\rmV \in \sR^{T \times H \times W \times 3}$, the VQ-VAE's encoder $E$ produces latent embeddings $\rmZ = E(\rmV) \in \sR^{T'\times H'\times W'\times d}$. 
Each embedding vector $\rvz \in \sR^d$ in $\rmZ$ is then passed through the vector quantizer $q$, which assigns it to the closest entry $\rvc \in \sR^d$ in the learned codebook embedding $\rmC \in \sR^{K \times d}$:
% The individual embeddings $\rvx \in \sR^d$ in each latent position are vector-quantized by nearest neighbor search over the learned codebook $\rmC=\sR^{K \times d}$,
% \lijun{is $\rvx$ the index or the post-quantization embedding? based on the wording in sec 3, embedding seems more suitable. Then I think we do not need $\rvc_\rvx$ and can just use $\rvx$ in the loss}
\begin{equation}
\vspace{-2mm}
    q(\rvz) = \rvc_i, \text{ where } i=\underset{j \in \{1,2,\cdots,K\}}{\argmin}\Vert \rvz - \rvc_j \Vert_2.
\end{equation}
To get discrete tokens, we drop the embedding dimension and represent $\rmZ$ by its indices $\rmX \in \{1, 2, \cdots, K\}^{T' \times H' \times W'}$. For decoding, embeddings of all image tokens are given as input to the decoder $D$ to reconstruct the input $\hat{\rmV} = D(\rmZ)$. 
%The standard version of the VQVAE is trained with the following loss optimizing jointly the encoder, decoder and codebook:
% \begin{equation}
% \mathcal{L}_{\text{vqvae}} = \| \rmV - \hat{\rmV} \|_2^2 + \| \text{sg}(\rmZ) - \rvc_\rmZ \|_2^2 + \beta \| \rmZ - \text{sg}(\rvc_\rmZ) \|_2^2 
% \end{equation}
% where $\beta$ is a hyperparameter balancing the codebook and encoder updates, and $\rvc_\rmZ$ denotes the concatenation of all quantized vectors. The straight-through gradient estimation is used, denoted by the stop gradient operation $\text{sg}$, to copy the gradients directly from the decoder to the encoder.
Following VQ-VAE, VQGAN~\citep{esser2021taming} introduces an adversarial loss and feature-level perceptual losses to enhance the image quality.

Video tokenization is more challenging and VQGAN has been adapted to meet this purpose~\citep{ge2022long,villegas2022phenaki,yu2022magvit}. The state of the art in video tokenization is MAGVIT~\citep{yu2022magvit}, which introduces a better 3D architecture, an inflation technique for initialization using image pre-training, and robust training losses. With MAGVIT, the LMs achieve leading generation quality across multiple video benchmarks. However, MAGVIT struggles to tokenize images and often results in noticeable flickering in longer videos.
%In the video domain, the encoder and decoder can be implemented using 3D CNNs 

% In this work we build upon the state-of-the-art video tokenizer MAGVIT from \cite{yu2022magvit}. MAGVIT introduces several innovations to video tokenization, obtaining high reconstruction quality and a discrete latent space that produces state-of-the-art video generation when coupled with a masked language model.
% Introduce the MAGVIT in more details.

% Introduce VQ-VAE and follow-up works.

%% file: sections/method.tex
%
%The complexity of modeling spatial-temporal dynamics motivated our focus on video tokenization rather than on image tokenization. 

We introduce a new \textbf{video tokenizer} designed to map the spatial-temporal dynamics from a visual scene into compact discrete tokens suitable for language models. 
\cl{Compared with image generation, video generation still faces substantial challenges in generating consistent and realistic motion. We are interested in exploring the capabilities of language models in tackling this unsolved challenge. Therefore, this paper focuses on a video tokenizer that can effectively represent video for generative modeling.}
Our approach builds upon the state-of-the-art video tokenizer, MAGVIT, as detailed in \citet{yu2022magvit}. 
This section highlights two new designs: a lookup-free quantizer and a collection of enhancements to the tokenizer model.
\vspace{-2mm}
\subsection{Lookup-Free Quantizer}
\vspace{-2mm}

Although the community has made great progress in developing VQ-VAEs, the relationship between improvements in the reconstruction quality and subsequent generation quality is still not well understood.
A common misconception is that improving reconstruction equates to improving the generation of the language model.
% For example, enlarging the VQ vocabulary could result in better reconstruction in \cref{fig:curve}, but it does not translate to an improvement of generation in \cref{fig:curve_gen} in many cases~\citep{yu2022scaling}.
For example, enlarging the vocabulary can improve reconstruction quality. However, such improvement only extends to generation when the vocabulary size is small, and a very large vocabulary can actually hurt the performance of the language model. 

\begin{wrapfigure}{r}{0.5\textwidth}
\vspace{-4mm}
    \centering
    % \begin{subfigure}{0.43\linewidth}
    \includegraphics[width=\linewidth,trim={0 0 0 0},clip]{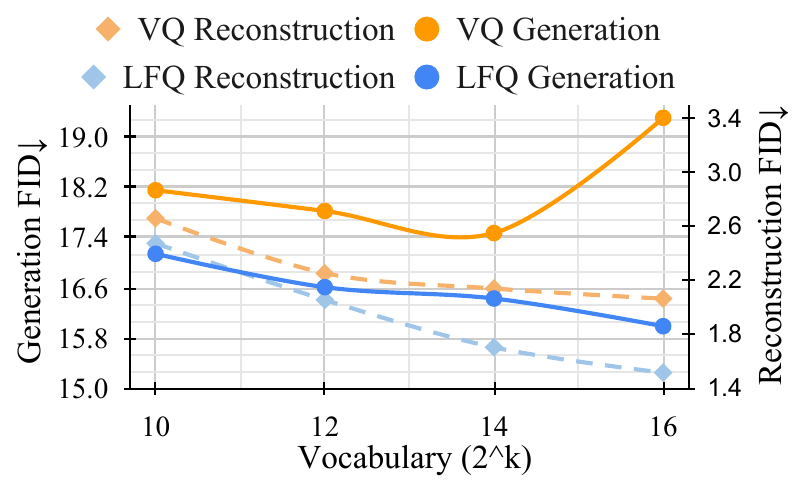}
    % \caption{Reconstruction quality.}
    % \label{fig:curve_recons}
    % \end{subfigure}
    % \begin{subfigure}{0.5\linewidth}
    % \includegraphics[width=\linewidth,trim={0 0 0 1cm},clip]{gen_curve.pdf}
    % \caption{Generation quality.}
    % \label{fig:curve_gen}
    % \end{subfigure}
    \vspace{-6mm}
    \caption{\textbf{Reconstruction and generation quality curves} in FID on ImageNet when scaling the tokenizer's vocabulary size with Vector Quantization (VQ) and Lookup-Free Quantization (LFQ). 
    Comparison is done at 128$\times$128 resolution using an MLM with 306-372M parameters.
    %  using a tokenizer with 52M and
    % \lu{Size of LM and tokenizer}
    % \lijun{todo}  \lu{change the color green to a warm color.}
    % The area of the dots indicate the parameter size for the tokenizers in (a) and language models in (b). VQ uses fixed 256 embedding dimensions, while VQ (lower embed) uses $2^{18-k}$ dimensions.
    }
    \label{fig:motivation}
    \vspace{-2mm}
\end{wrapfigure}

As illustrated by the dashed curves in \cref{fig:motivation}, the reconstruction FID, indicated by the right $y$-axis (where a lower value is better), improves as the vocabulary size (the $x$-axis) increases. The orange solid curve in \cref{fig:motivation} represents the LM's generation quality (the left $y$-axis). The generation FID initially improves but deteriorates for larger vocabulary. This may shed light on why the vocabulary size of most language models for visual generation is around 1-8k~\citep{esser2021taming,villegas2022phenaki}, which is significantly smaller than the size of natural language vocabulary, \ie over 200k.

A simple trick for training a larger codebook involves decreasing the code embedding dimension when increasing the vocabulary size~\citep{yu2021vector}.
This trick captures the intuition of limiting the representational capacity of individual tokens, which in turn facilitates learning over the distribution of a large vocabulary.

% ~\citep{rakhimov2021latent,le2021ccvs}

%, \ie increasing $K$ while reducing  in the vocabulary $\rmC^{K\times d}$.

\vspace{-4mm}
\paragraph{Lookup-Free Quantization (\quantizername{}).}

Motivated by the above observation, we reduce the VQ-VAE codebook's embedding dimension to zero. Formally, the codebook $\rmC \in \sR^{K \times d}$ is replaced with an integer set $\sC$ where $|\sC| = K$.
% reduced to an integer set $\sC = \{1, 2, \cdots, K\}$. 
Recall that in VQ-VAE models, the quantizer must look up all $K$ $d$-dimensional embeddings in the codebook, where $d$ is typically $256$, when computing the closest codebook entry to the encoder output. This new design eliminates the need for such embedding lookup entirely hence we call it \emph{lookup-free quantization (\quantizername{})}.
We found that \quantizername{} can grow the vocabulary size in a way benefiting the generation quality of language models. As shown by the blue curves in \cref{fig:motivation}, both reconstruction and generation consistently improves as the vocabulary size increases -- a property not observed in current VQ-VAE methods.

While various \quantizername{} methods are available, this paper discusses a straightforward variant that assumes independent codebook dimensions and binary latents. 
Specifically, the latent space of \quantizername{} is decomposed as the Cartesian product of single-dimensional variables, as $\sC = \bigtimes_{i=1}^{\log_2{K}}C_i$.
%Lookup-free quantization can scale to larger vocabulary in a way that benefits the generation of language model. We set the vocabulary size to $2^{18}$ comparable to that of PaLM to leverage the instrcture and learning receipes built for LLMs.
Given a feature vector $\rvz \in \sR^{\log_2{K}}$, each dimension of the quantized representation $q(\rvz)$ is obtained from:
\begin{equation}
% \vspace{-1mm}
    q(\ervz_i) = C_{i, j}, \text{ where } j = \argmin_k \|\ervz_i - C_{i, k}\|,
\end{equation}
%Depending on the values in $C_i$, we have various ways to compute the $\argmin$. However,
where $C_{i, j}$ is the $j$-th value in $C_i$.
With $C_i = \{-1, 1\}$, the $\argmin$ can be computed by the sign function as 
% We call it \emph{lookup-free binary quantization (LBQ)}:
\begin{equation}
% \vspace{-1mm}
    q(\ervz_i) = \sign(\ervz_i) = -\1\{\ervz_i \le 0\} + \1\{\ervz_i > 0\}.
\end{equation}
% which can be easily done via hashing or rounding according to the values in $C_i$.
% where $\sC_i$ is the predefined ordered list of integers for the $i$-th dimension.
% % The effective size of the vitual codebook becomes $|\sC| = \prod_{i=1}^N |C_i|$.
% To get the token index in $\sC$, we have
With \quantizername{}, the token index for $q(\rvz)$ is given by:
\begin{equation}
\vspace{-1mm}
\mathit{Index}(\rvz) = \sum_{i=1}^{\log_2{K}} \argmin_k \|\ervz_i - C_{i, k}\| \prod_{b=0}^{i-1} |C_b| = \sum_{i=1}^{\log_2{K}} 2^{i-1} \1\{\ervz_i > 0\}, \label{eq:tokenid}
\end{equation}
where $|C_0|=1$ sets the virtual basis.

% where $N=\log{K}$ and all $\sC_i = \{-1, 1\}$. 
% As a result, we have
% \begin{equation}
%     \mathit{Id}(\rvx) = \sum_{i=1}^N 2^{i-1} \1\{x_i > 0\} \label{eq:tokenid}
% \end{equation}
% The number of tokens  $N=\log_2{K}$ 
% In addition, $|\sC| = 2^N$ and $B=N$.
% The overall space and time complexity is just $O(B)$.

We add an entropy penalty during training to encourage codebook utilization:
% Some VQ models, \eg in \citep{chang2022maskgit}, involves an entropy penalty~\citep{jansen2020coincidence} during training to encourage the codebook utilization.
\begin{align}
    \gL_\mathit{entropy} &= \E[H(q(\rvz))] - H[\E(q(\rvz))].
    % = \E \big[\sum_{i=1}^{\log_2{K}} H(\ervq (\ervx_i))\big] - H[\E(\rvq(\rvx))]
    % H(\rvq(\rvx)) &= \sum_{i=1}^{\log_2{K}} H(\ervq_i (\ervx_i)) 
    \label{eq:entropy}
\end{align}
This penalty is inspired by a similar loss used in image VQGAN model~\citep{chang2022maskgit}, which is also found in entropy-based clustering~\citep{jansen2020coincidence}. In \quantizername{}, given the independence among dimensions, we rewrite $H(q(\rvz)) = \sum_{i=1}^{\log_2{K}} H(q (\ervz_i))$ .
The $H[\E(q(\rvz))]$ term can be approximated with sub-groups of dimensions for $K>2^{18}$ where direct estimation is memory bound. 

% It prevents unhealthy initial states of the model where some bits are never activated.
% leads to a healthy initial state of the model where all bits are utilized.
% With the decomposed latent space in \quantizername{}, the first term can be directly decomposed and computed efficiently as $H(\rvq(\rvx)) = \sum_{i=1}^N H(\ervq_i (\ervx_i))$.

We note that there are various other variants of \quantizername{}, \eg, opting for the multivariant over the binary codebook $C_i$ or employing other quantization techniques such as \citet{agustsson2019generative}.
%or using mutivaint codebook instead of the binary code. 
As the first paper to introduce this concept, we focus on the simplest form with independent binary dimensions, which shows promising improvements. Other \quantizername{} methods merit further research. 
% as $H[\E(\rvq(\rvx)] \approx \E_\mathit{\rvg}\big[H[\E(\rvq_\rvg(\rvx_\rvg)] \big]$.
    % , \rvg \in \{(1, 2, ..., n), (n+1, n+2, ..., 2n), ... (N-n+1, N-n+2, ..., N), (1, n+1, ..., N-n+1), ..., (n, 2n, ..., N)  \}
% The tokens produced by VQ are meaningless integers over a categorical distribution defined by the embeddings in the codebook, which downstream models usually do not have access to.
% On the other hand, LQ tokens represent the embedding by themselves and have a well defined distance metric.

% Generative transformers~\citep{ramesh2021zero,esser2021taming,chang2022maskgit,yu2022magvit} ususally adopt a two-stage framework including a VQ-based tokenizer and a transformer operating in the token space.

% Unless specially designed, the transformer is agnostic to the VQ codebook but only tries to fit the latent distribution of meaningless integers.

% usually operate in the discrete latent space learned by VQ.
% One approach we have found to work best is binary quantization.

\cl{In addition to the entropy penalty (\cref{eq:entropy}), an \quantizername{}-based tokenizer is trained using the standard combination of \emph{reconstruction},  \emph{GAN}, \emph{perceptual}, and \emph{commitment} losses~\citep{esser2021taming}, excluding the inapplicable codebook loss. Following \cite{yu2022magvit}, we use LeCAM regularization~\citep{tseng2021regularizing} for improved stability. 
% For the latents, besides the entropy penalty~(\cref{eq:entropy}), we also use the \emph{commitment} loss~\citep{van2017neural},  $\mathcal{L}_{comm} = \|\rvz - \text{stopgrad}(q(\rvz))\|_2^2$.
}

\subsection{Visual Tokenizer Model Improvement}\label{sec:tokenizer_improvement}
\begin{figure}[tp]
\vspace{-6mm}
    \centering
    \begin{subfigure}{1\textwidth} % no visible
    \refstepcounter{subfigure}\label{fig:arch:a}
    \refstepcounter{subfigure}\label{fig:arch:b}
    \refstepcounter{subfigure}\label{fig:arch:c}
    \end{subfigure}%
    \includegraphics[width=0.95\linewidth,trim={0 0 0 0},clip]{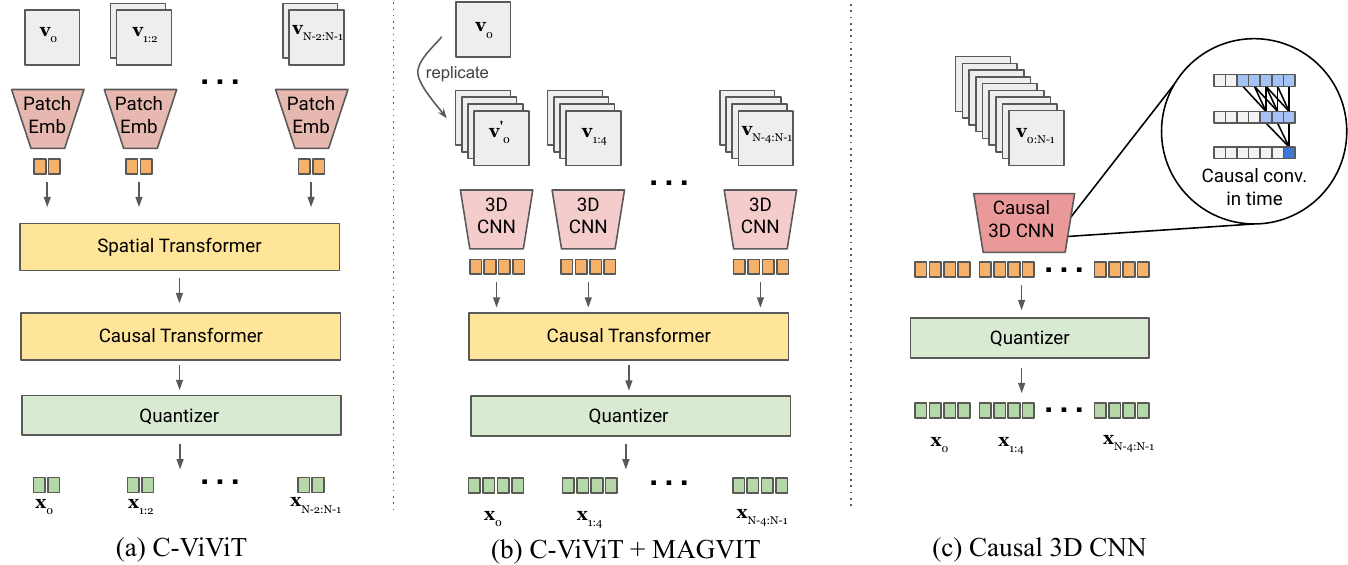}
    \vspace{-2mm}
    \caption{\textbf{Causal tokenizer architecture comparison}.
    % \textbf{(a)} C-ViViT. \textbf{(b)} Hybrid of C-ViViT and MAGVIT, where the 3D CNN processes blocks of 4 frames, connected by a causal transformer.
    % \textbf{(c)} \modelname{}: causal 3D CNN using causal convolutions in the temporal dimension. 
    The decoders, which are omitted from the figure, employ an architecture that is symmetric to the encoder.
    \cl{\scriptsize See detailed architecture diagram in the Appendix.}
    }
    \label{fig:arch}
    \vspace{-6mm}
\end{figure}

%The devil is in the details.

% \lu{could this be a story: we need to tokenize image and video into the same codebook -> need to tokenize image seperately -> need to get causal design for image -> pros and cons of such design.}

% To fully leverage the enhanced modeling capability brought by the new quantizer, we introduce a new tokenizer model design that unifies the tokenization of images and variable length videos, with causal temporal dependency enforced.

% new designs for the encoder, decoder, and discriminator modules in the tokenizer.
% In particular, the new model 

% \lijun{causal 3d cnn figure}

\vspace{-2mm}
\paragraph{Joint image-video tokenization.}
A desirable feature of visual tokenization is the capability to tokenize images and videos using a shared codebook. However, the MAGVIT tokenizer, which utilizes the 3D CNN, faces challenges in tokenizing images due to the temporal receptive field. %While replicating an image multiple times to form a static video is possible, this approach introduces redundant code and can potentially alter the training data distribution, resulting in reduced motion.

To build a joint image-video tokenizer, a new design is needed. We begin our discussion by revisiting an existing method C-ViViT~\citep{villegas2022phenaki}. As depicted in \cref{fig:arch:a}, C-ViViT employs full spatial transformer blocks combined with causal temporal transformer blocks. This approach performs reasonably well but has two drawbacks. First, unlike CNNs, the positional embeddings makes it difficult to tokenize spatial resolutions that were not seen during training. Second, empirically we found that 3D CNNs perform better than spatial transformer and produce tokens with better spatial causality of the corresponding patch.
% the self-attention creates a global dependency among tokens. Though such dependency also exists in CNN, it exhibits more globally in the self-attention across all tokens. Hence, a change to a local pixel patch can influence many tokens that are further away,  posing challenges to video editing.
% \jose{doesnt this happen in 3dcnn with enough receptive field as well?}\lijun{I think it remains local spatially, e.g., for an encoder with 16 conv layers on 128x128} 

To tackle these drawbacks, we explore two plausible designs. \cref{fig:arch:b} combines C-ViViT and MAGVIT. Assuming a temporal compression ratio of 4, a 3D CNN processes blocks of 4 frames followed by a causal transformer. In \cref{fig:arch:c}, we use the temporally causal 3D convolution to replace the regular 3D CNN. Specifically, the temporal padding scheme for a regular 3D convolution layer with kernel size $(k_t, k_h, k_w)$ includes $\lfloor\frac{k_t-1}{2}\rfloor$ frames before and $\lfloor\frac{k_t}{2}\rfloor$ frames after the input frames.
In contrast, a causal 3D convolution layer pads with $k_t - 1$ frames before the input and nothing after, so that the output for each frame only depends on the previous frames.
% In this way, at all layers of a deep network, the feature of each frame only depends on the previous frames.
In consequence, the first frame is always independent of other frames, allowing the model to tokenize single images.

Temporal convolutional subsampling with stride $s$ is sufficient for $s\times$ down-sampling by mapping $1 + s \times t$ frames into $1 + t$.
After a regular $s \times$ up-sampling, we drop the first $s - 1$ resulting frames, which maps  $1 + t$ frames into $1 + s \times t$ and allows for the tokenization of a single image. \cref{tab:ablation_arch} empirically compares the designs in \cref{fig:arch}, and we find that the causal 3D CNN performs the best.

\vspace{-5mm}
\paragraph{Architecture modifications.}
% \lijun{we have ablation results for each design}
In addition to using causal 3D CNN layers, we made several other architectural modifications to improve upon the MAGVIT model. First, we change the encoder downsamplers from average pooling into strided convolutions to leverage learned kernels, and replace the decoder upsamplers from
% First, the average pooling layers in the encoder are replaced with strided convolutions to leverage learned kernels. Second, for the decoder's upsampling layers, the previous method of 
nearest resizing followed by convolution with a depth-to-space operator.
Second, we defer the temporal downsampling from the first few encoder blocks to the last ones.
In addition, the downsampling layer in the discriminator now utilizes 3D blur pooling~\citep{zhang2019making} to encourage shift invariance. 
Finally, we add one adaptive group normalization layer before the residual blocks at each resolution in the decoder to pass in the quantized latents as the control signal following StyleGAN~\citep{karras2019style}.
\cref{tab:ablation_imagenet,tab:ablation_ucf101} empirically verify these designs.
%We also double the number of residual blocks at each resolution to increase the model parameters.

% Besides the pure convolution layers, we revisit the designs of other model components to maximize the performance.
% In particular, we change the average pooling layers in the encoder into strided convolutions to leverage learned kernels.

%\lu{any change in inflation method? This is new in MAGVIT and we can }

\vspace{-5mm}
\paragraph{Token factorization for efficient prediction.}
The output tokens can be fed into language models to generate videos. To assist smaller transformers predicting in a large vocabulary, we can factorize the \quantizername{} token's latent space into equal subspaces. For instance, rather than predicting using a codebook of size $2^{18}$, we can predict in two concatenated codebooks, each of size $2^9$. We embed each subspace token separately and use their embedding summation as the token embedding for the transformer input. \cl{We find  it beneficial to use weight tying~\citep{press2017using}, a common technique in language modeling, which involves sharing the weights between the embedding and softmax layers.
For the output layer with a factorized vocabulary}, we use the embedding matrix for each subspace to obtain logits with seperate prediction heads. 

%% file: sections/experiments.tex
\begin{figure}[tp]
\vspace{-6mm}
    \centering
    \includegraphics[width=\linewidth]{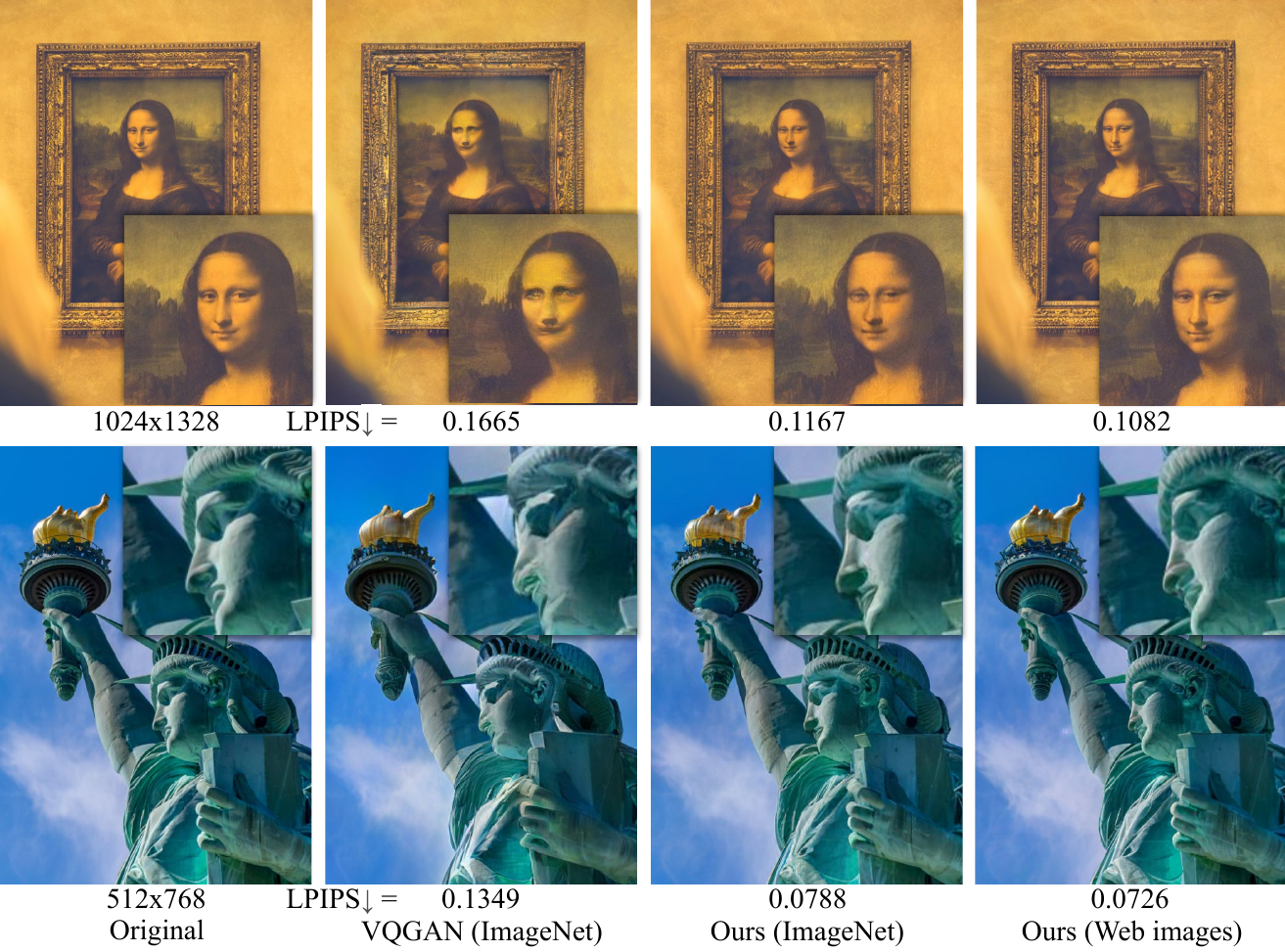}
    \vspace{-4mm}
    \caption{\textbf{Image reconstruction samples with different tokenizers}. We compare the VQGAN used in MaskGIT~\citep{chang2022maskgit} with two of our models trained on ImageNet and web images~\citep{chen2022pali}. \mytiny{Original images are by Eric TERRADE and Barth Bailey on Unsplash}.}
    % https://unsplash.com/photos/0WQOCx1g8hw
    % https://unsplash.com/photos/s0di82cRiUQ
    \label{fig:reconstruction}
    \vspace{-4mm}
\end{figure}

This section empirically verifies the proposed tokenizer across three distinct tasks: video and image generation, video compression, and action recognition.
\cref{fig:reconstruction} visually compares the reconstruction quality of our tokenizer with prior works.
More qualitative samples are shown at \webpage{}.

\vspace{-2mm}
\subsection{Experimental Setups}
\vspace{-2mm}
\paragraph{Datasets.} 
We use Kinetics-600 (K600)~\citep{carreira2018short} and UCF-101~\citep{soomro2012ucf101} for video generation experiments, along with ImageNet~\citep{deng2009imagenet} for image generaton.
In addition, MCL-JCV~\citep{wang2016mcl} is used as the testbed for video compression, with Kinetics-400 (K400)~\citep{kay2017kinetics} and SSv2~\citep{goyal2017something} for video understanding. 
% For compression evaluation we used the dataset: MCL-JCV~\citep{wang2016mcl}.
\vspace{-4mm}
\paragraph{Implementation details}
We follow the tokenizer training setting and hyperparameters in~\citep{yu2022magvit}, unless stated otherwise. \quantizername{} is used, which eliminates the codebook embedding, to increase the default codebook size to $K=2^{18}$. 
%that is comparable to that of PaLM 2~\citep{googlepalm2}.
The weight of $\gL_\mathit{entropy}$ follows an annealing schedule with a $3\times$ higher starting point and linearly decays to a fixed value of $0.1$ within 2k steps.
%We also $5\times$ upweight the $\ell_2$ reconstruction loss to encourage better fidelity.
We defer details regarding the evaluation setup of each subsection to the Appendix.
% \lu{talk about factorization of two codebooks}
% The same MLM as in \citep{yu2022magvit} is used.
% \citet{yu2022magvit} inflates a 2D convolution layers into a regular 3D one
% by filling in the temporally central slice of a zero-filled kernel.
% For the causal 3D layers, we adapt the inflation strategy to fill in the temporally last slice to correspond to the causal padding scheme.
% In addition, we disable the inflation for the discriminator and train it from scratch for better stability.
% We train our model using the infrastructure \url{https://github.com/google-research/magvit}
% \lu{add all the tables; it's fine some numbers are missing}.

\vspace{-2mm}
\subsection{Visual Generation}\label{sec:gen}
\vspace{-2mm}
The masked language model (MLM)~\citep{devlin2019bert} is used in image and video generation. To verify the tokenizer, we employ the same MLM transformers in MAGVIT~\citep{yu2022magvit}. \cl{We select the MLM due to its competitive performance on benchmark datasets~\citep{yu2022magvit,lezama2023discrete}. In the Appendix, we also show that an autoregressive language model (AR-LM) coupled with the proposed tokenizer outperforms the prior work MAGVIT.}
As we use a smaller MLM ($\sim$300M parameters) with a large codebook ($2^{18}\approx$262K), the token factorization as discussed in \cref{sec:tokenizer_improvement} is applied using two heads with each predicting from a codebook of size $2^9$.

\input{tables/generation_v}

\begin{figure}[tp]
\vspace{-4mm}
    \centering
    \includegraphics[width=\linewidth,trim={0 0 0 0},clip]{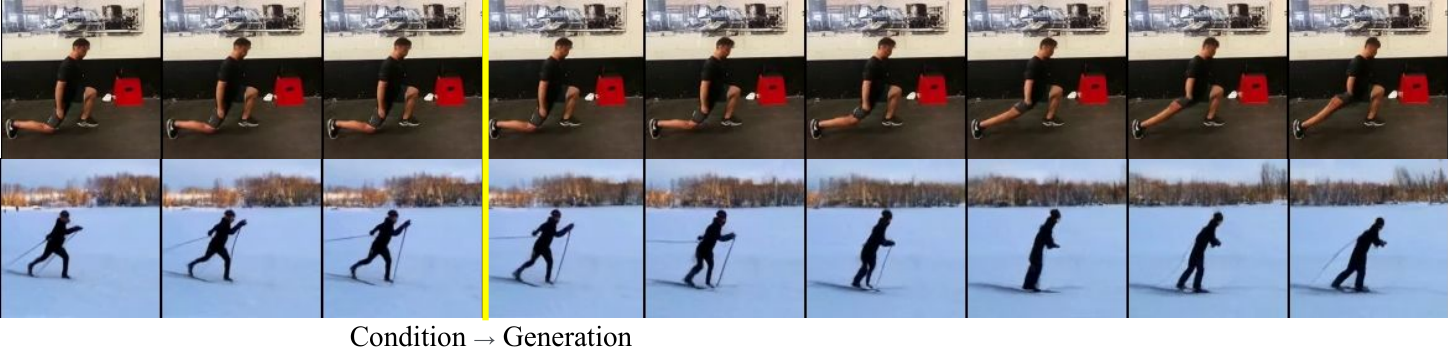}
    \vspace{-7mm}
    \caption{\textbf{Frame prediction samples on Kinetics-600}.}
    \label{fig:k600}
    \vspace{-2mm}
\end{figure}

\vspace{-4mm}
\paragraph{Video generation.}
We consider two standard video benchmarks,
UCF-101 for class-conditional generation and K600 for frame prediction with 5-frame condition.
FVD~\citep{unterthiner2018towards} is used as our primary evaluation metric. 
% We trained the video tokenizer on Kinetics-600 training set for 190 epochs with batch size 256, then trained the MLM transformer following \cite{yu2022magvit} with token factorization for 360 epochs with batch size 256. 
% Further implementation details are described in the appendix. 
\cref{tab:gen_k600} shows that our model surpasses all prior arts in both benchmarks.
Specifically, it outperforms the previous best model MAGVIT by a large margin, while using the same MLM transformer backbone. 
\cl{In addition, it significantly outperforms the non-causal baseline on frame prediction, highlighting the contribution of the causal tokenizer.}
These results demonstrate the essential role of a good visual tokenizer in enabling LMs to generate high-quality videos.
\cref{fig:k600} shows qualitative samples from the model.

\input{tables/generation_i}

\vspace{-4mm}
\paragraph{Image generation on ImageNet.}
We evaluate \modelname{} on image generation under the standard ImageNet class-conditional setting.
We present results for resolution 512$\times$512 in \cref{tab:gen_in512} and refer to the Appendix for 256$\times$256 results. FID~\citep{heusel2017gans} and Inception Score (IS)~\citep{salimans2016improved} are used as evaluation metrics. Our model surpasses the best performing diffusion models both in sampling quality (FID and IS) and inference-time efficiency (sampling steps). 

\begin{figure}[tp]
% \vspace{-4mm}
    \centering
    \includegraphics[width=\linewidth]{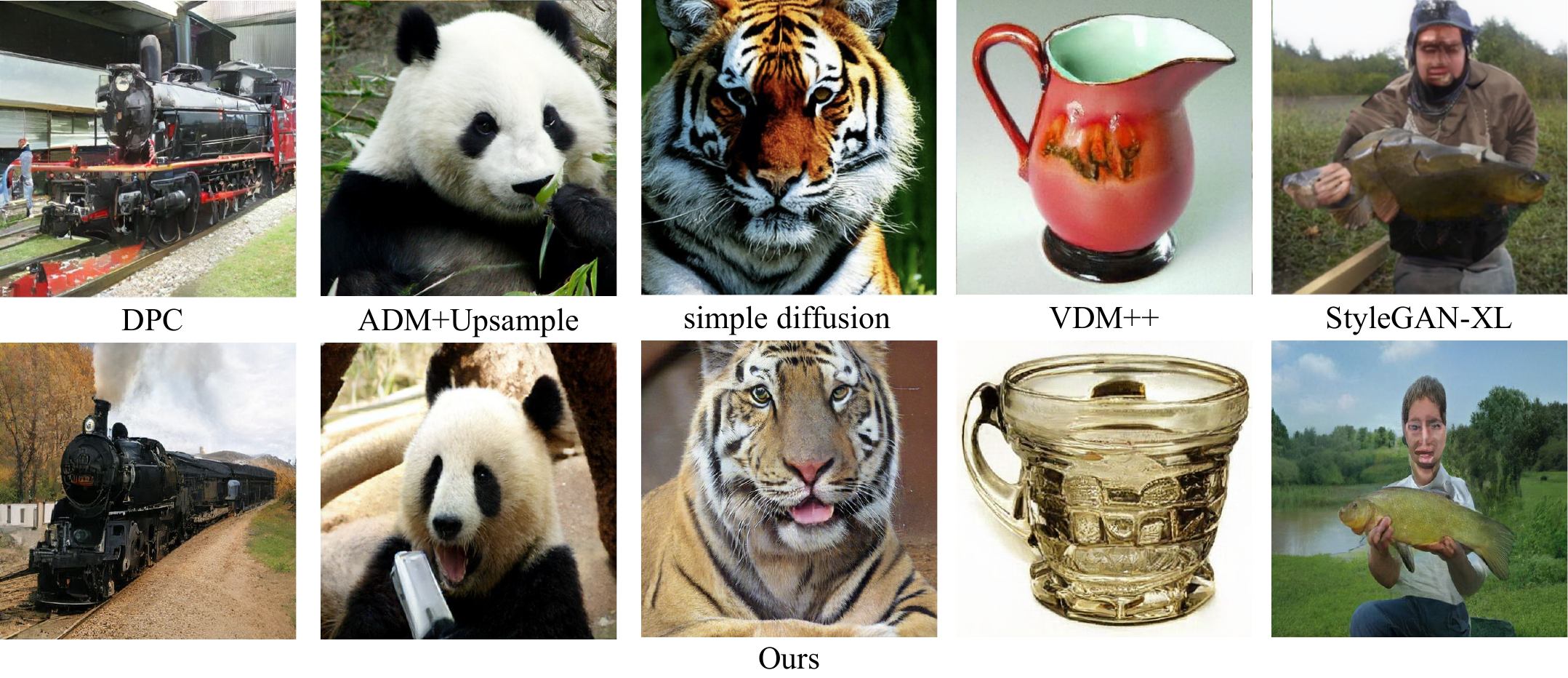}
    \vspace{-7mm}
    \caption{\textbf{Class-conditional generation samples on ImageNet 512$\times$512}. We compare with each of the previous works with a random sample from the same image class.}
    \label{fig:imagenet}
    \vspace{-6mm}
\end{figure}

It is worth noting that all the models compared are trained using the same ImageNet training data, with a comparable model size and training budget. Therefore, the performance primarily evaluates the model's capabilities. The masked language model, equipped with our tokenizer, exhibits a notable improvement in FID over the best diffusion model baseline at 512$\times$512 (FID=1.91 \vs 2.65, 28\%$\downarrow$). While this margin narrows at 256$\times$256 resolution, the MLM uses a 50\% reduced model size and needs much fewer decoding steps (\eg, 64 \vs 250) to get the image generation quality.
% It is noteworthy that all the compared models are trained on the same training data, using comparable modelsize, and training budget. So the performance reflects is a test of the model itself. Our model shows significant FID improvement over the best diffusion on 512X512 (FID=1.91 vs. ).
% Albeit the margin is smaller, our model uses 50\% smaller model and much smaller decoding step to generate the image.
% \jose{128x128?}  \lijun{no 128}
% We trained the tokenizer on the ImageNet training set for 270 epochs using a batch size of 256. With this tokenizer we trained a Masked Language Model following \cite{chang2022maskgit}, with the token factorization described in Section~\ref{sec:tokenizer_improvement}, for 1080 epochs and with batch size 1024. We refer to the appendix for further implementation details. \jose{describe augmentations?} 
Qualitative samples in comparison with other models are shown in \cref{fig:imagenet}.

% \cref{} shows the ImageNet generation results. Our model is compared with   
% 256: appendix

% \subsection{Video Generation and Future Frame Generation}
% \vspace{-4mm}

\setlength\intextsep{20pt}
\begin{wrapfigure}{r}{0.45\textwidth}
\vspace{-2mm}
    \centering
    %\begin{subfigure}{.48\linewidth}
    % Figure generated by this colab:
    % https://colab.corp.google.com/drive/1jpXXVPyztiyUPFxrfkdpubZ0NmEJAAW3?resourcekey=0-VKb_gdWn7Gp4Sy4e5MRdNg
    \includegraphics[width=\linewidth]{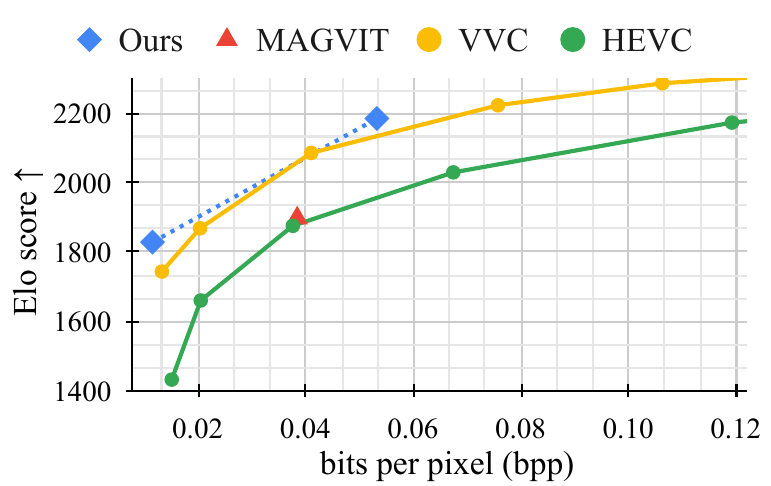}
    %\end{subfigure}
    %\begin{subfigure}{0.48\linewidth}
    %\includegraphics[width=\linewidth]{figures/psnr-codec-curves.pdf}
    %\end{subfigure}
    \vspace{-6mm}
    \caption{\textbf{Video compression rater study}.
    % In a subjective rater study assessing compression quality, our model is preferred to other video models including MAGVIT, HEVC (H.265), and VVC (H.266). We calculate Elo scores based on pairwise preferences to quantify the relative visual quality between the models (higher is better).
    }
    \label{fig:compression}
    \vspace{-28mm}
\end{wrapfigure}
\vspace{-8mm}
\leavevmode\subsection{Video Compression}

We conduct a subjective rater study to assess the compression quality of \modelname{}. The study is conducted on the 30 videos of the MCL-JCV dataset, resized to a resolution of 640$\times$360.
Sixteen raters are engaged, each providing responses to an average of roughly 800 pairwise-preference questions.
% conducted using 16 raters, with an average of approximately 800 pairwise-preference questions per rater.
% The questions were presented with an interface that parallels the one used for the Challenge on Learned Image Compression (http://compression.cc/), extended to comparing videos. Raters were not allowed to pause the videos.
% The study was conducted on the 24 videos of the MCL-JCV dataset (https://mcl.usc.edu/mcl-jcv-dataset/), scaled down to a resolution of 640x360 pixels.
\newpage

We calculate Elo scores~\citep{elo1978rating} based on pairwise preferences to quantify the relative visual quality between the models. The study compares our model with MAGVIT as well as the current video compression standard HEVC (H.265) video codec~\citep{sullivan2012overview} and the next-generation codec VVC (H.266)~\citep{vvc}. As shown in \cref{fig:compression}, raters prefer our model to the compared methods at multiple bit rates.

\input{tables/compression}
%\lu{add a sentence to give a bit more context on HEVC abd VCC}.
% Here we will add a figure comparing MAGVIT, HEVC, VCC compression quality in user studies.
We also compare the compression quality using common distortion metrics (LPIPS, PSNR, and MS-SSIM). 
\cl{\cref{tab:compression} compares at 0.0384 bpp, the bit rate of MAGVIT, with full curves in the Appendix}.
The results show that our model outperforms MAGVIT on all metrics, and it outperforms all methods on LPIPS, a metric which correlates more closely with subjective quality assessments.
%\cl{At the same bit rate, standard codecs 
% based on block-level Fourier-related transformations may capture local details more accurately than neural models but introduce block artifacts, which can result in higher PSNR and MS-SSIM due to their lack of realistic-ness measurement~\citep{agustsson2019generative}.}
% According to \citet{agustsson2019generative}, PSNR and MS-SSIM focus on local details but cannot measure realistic-ness, where
%
\cl{
At equal bit rates, standard codecs may render local details
 more accurately than neural models but  also introduce block
 artifacts,  detrimental to  perceptual quality yet not captured by PSNR and MS-SSIM~\citep{agustsson2019generative}.
}
%\cl{Future research is needed to improve the efficiency of the tokenizer to match that of standard codecs.}
% \cl{Further research is needed to implement a CPU version of the tokenizer matching the compute efficiency of standard codecs.}
\cl{Despite promising results with TPUs, further research is needed to adapt our model to run efficiently on CPUs like standard codecs.}

%Since PSNR and MS-SSIM correlate relatively poorly with subjective assessments of visual quality, it is not unexpected that our method scores worse on these distortion metrics than HEVC and VVC.

% \newfloatcommand{capbtabbox}{table}[][\FBwidth]

% \begin{figure}
% \begin{floatrow}
% \ffigbox{%
%   \includegraphics[width=\linewidth]{figures/elo-codec-curves.pdf}%
% }{%
%   \caption{In a human rater study of compression quality, our model is preferred to other video models including MAGVIT, HEVC (H.265), and VVC (H.266), as shown by relative Elo scores.}
%   }
% \capbtabbox{%
%     \begin{tabular}{lcccc}
%     \toprule
%       & PSNR$\uparrow$ & MS-SSIM$\uparrow$ & LPIPS$\downarrow$\\
%     \midrule
%     HEVC & 30.15 & 0.94 & 0.20 \\ 
%     VVC & 32.72 & 0.97 & 0.15\\ 
%     MAGVIT\\ 
%     Ours\\
%     \bottomrule
%     \end{tabular}

% }{%
%   \caption{Quantitative metrics averaged over the MCL-JCV video dataset targeting an average bpp of 0.0391.}%
% }
% \end{floatrow}
% \end{figure}

% \vspace{-2mm}
\input{tables/understanding}
\subsection{Video Understanding}
\vspace{-2mm}
In this subsection, we assess the tokenizer's capability to learn a video understanding model for action recognition. Two setups are examined: (1) using tokens as prediction targets for the transformer's output, and (2) using tokens as the input to the transformer. For the former setup, we use a similar architecture following the BEVT~\citep{wang2022bevt} pre-training. For the tokens as inputs, to work with the ViViT backbone~\citep{arnab2021vivit}, we detokenize the tokens to pixels before feeding them to \cl{frozen} ViViT transformers \cl{trained on raw pixels}.

%A good tokenizer would excel in both settings.

\cref{tab:understanding} shows that \modelname{} outperforms the previous best MAGVIT in these evaluations. Specifically, when using the decoded tokens as input, the performance approaches that of the model trained with ground-truth pixels using the same ViViT backbone. While these numbers are still worse than the state-of-the-art in action recognition, they represent solid improvements credited to the new tokenizer.
% Here we will compare BEVT training using different tokenizers other than the proposed one.

\vspace{-2mm}
\subsection{Ablation Study}
\vspace{-2mm}
In \cref{fig:motivation}, we have ablated \quantizername{} \vs VQ and the vocabulary size.
In \cref{tab:ablation}, we validate the key designs proposed in \cref{sec:tokenizer_improvement}. Specifically, \cref{tab:ablation_arch} compares the architecture illustrated in \cref{fig:arch}; \cref{tab:ablation_imagenet} and \cref{tab:ablation_ucf101} verify the \quantizername{} and other improvements on ImageNet and UCF-101, respectively.
% vocabulary size: figure 1
% vq vs lq: figure 1
% table 4
% group transformer
% sq: appendix

\input{tables/ablation}

%% file: tables/generation_v.tex
\begin{table}[tp]
\centering
\vspace{-6mm}
\caption{\textbf{Video generation results}: frame prediction on Kinetics-600 and class-conditional generation on UCF-101. We adopt the evaluation protocol of MAGVIT.
% \emph{Method types}: AutoRegressive Language Model (ARLM), Masked Language Model (MLM), Generative Adversarial Network (GAN), continuous Diffusion (Diff.). 
% \emph{Latent types}: \quantizerfullname{} (\quantizername{}), Vector Quantization (VQ), Variational AutoEncoder (VAE), Discrete Cosine Transformation (DCT). 
% We report the bits per pixel (bpp) of the representation used in each latent-space model.
}
\label{tab:gen_k600}
\vspace{-2mm}
% \resizebox{\linewidth}{!}{%
\centering
\begin{tabular}{@{}l@{\hspace{5pt}}l@{\hspace{5pt}}c@{\hspace{5pt}}c@{\hspace{5pt}}c@{\hspace{3pt}}c@{}}
\toprule
Type & Method   & K600 FVD$\downarrow$  & UCF FVD$\downarrow$  & \#Params  & \#Steps  \\ \midrule
% \multicolumn{5}{l}{\textit{Pixel-Space Generative Models}} \\
% ARLM & Video Transformer~\citep{weissenborn2019scaling} & 170.0\mytiny{$\pm$5.0} & & 373M & 197k  \\
% GAN & DVD-GAN-FP~\citep{clark2019adversarial}  & 69.1\mytiny{$\pm$1.2} & & & 1                    \\
GAN & TrIVD-GAN-FP~\citep{luc2020transformation}  & 25.7\mytiny{$\pm$0.7} &  & & 1                \\
% Diffusion & RaMViD~\citep{hoppe2022diffusion}   &  17.5\mytiny{$\pm$1.1} & & 308M & 1000  \\
Diffusion & Video Diffusion~\citep{ho2022video}  & 16.2\mytiny{$\pm$0.3} & & 1.1B &   256        \\
Diffusion & RIN~\citep{jabri2023scalable} & 10.8 & & 411M & 1000 \\
\hdashline
% \multicolumn{6}{l}{\textit{Latent-Space Generative Models}} & Latent bpp\\
% ARLM + VQ & LVT~\citep{rakhimov2021latent} & 224.7 &  & 50M  & 16k  \\
% ARLM + VQ & CogVideo$^*$~\citep{hong2022cogvideo}  & 109.2 & 626 & 9.4B   & 64k     \\
% ARLM + VQ & CCVS~\citep{le2021ccvs}   & 55.0\mytiny{$\pm$1.0} & 386\mytiny{$\pm$18} & 361M   & 1024 \\
AR-LM + VQ & TATS~\citep{ge2022long}  &  & 332\mytiny{$\pm$18} & 321M & 1024 \\
MLM + VQ & Phenaki~\citep{villegas2022phenaki} & 36.4\mytiny{$\pm$0.2} & & 227M   & 48 \\
% ARLM + DCT & Transframer~\citep{nash2022transframer}  & 25.4  &  & 662M &  $<$4k      \\
% T MAGVIT-B-FP        & 24.5\mytiny{$\pm$0.9}     & 43.3\mytiny{$\pm$1.8} & 88M & 0.039 & 12 \\
MLM + VQ & MAGVIT~\citep{yu2022magvit}        & 9.9\mytiny{$\pm$0.3} & 76\mytiny{$\pm$2} & 306M   & 12  \\ 
\midrule
\cl{MLM + \quantizername{}} & \cl{non-causal baseline} & \cl{11.6\mytiny{$\pm$0.6}} & & \cl{307M} & \cl{12} \\
% \cl{AR-LM + \quantizername{}} &  & & \cl{?} & \cl{840M} & \cl{1280} \\
\multirow{2}{*}{MLM + \quantizername{}} & \multirow{2}{*}{\emph{\modelname{} (this paper)}} & 5.2\mytiny{$\pm$0.2} &  & \multirow{2}{*}{307M} & 12  \\
 & & \textbf{4.3\mytiny{$\pm$0.1}}  & \textbf{58\mytiny{$\pm$3}} & & 24  \\
\bottomrule
\end{tabular}
% \vspace{-2mm}
% }
\end{table}

%% file: tables/generation_i.tex
\begin{table}[tp]
\vspace{-2mm}
\caption{\textbf{Image generation results}: class-conditional generation on ImageNet 512$\times$512.
% \lu{we can remove GAN.} \lijun{recent diffusion papers all kept those, esp. StyleGAN-XL which is very strong}
Guidance indicates the classifier-free diffusion guidance~\citep{ho2021classifier}.
$^*$ indicates usage of extra training data.
We adopt the evaluation protocol and implementation of ADM.
}
\label{tab:gen_in512}
\vspace{-2mm}
\resizebox{\linewidth}{!}{%
\begin{tabular}{@{}l@{\hspace{5pt}}l@{\hspace{4pt}}c@{\hspace{2pt}}c@{\hspace{3pt}}c@{\hspace{2pt}}c@{\hspace{2pt}}c@{\hspace{2pt}}c@{}}
\toprule
\multirow{2}{*}{Type}  & \multirow{2}{*}{Method}  & \multicolumn{2}{c}{w/o guidance} & \multicolumn{2}{c}{w/ guidance}  & \multirow{2}{*}{\#Params}  & \multirow{2}{*}{\#Steps}  \\
& & FID$\downarrow$ & IS$\uparrow$ & FID$\downarrow$ & IS$\uparrow$ \\ \midrule
% \multicolumn{6}{l}{\textbf{512 $\times$ 512 resolution}} \\
% \multicolumn{6}{l}{\textit{Pixel-Space Generative Models}} \\
% ADM & & 23.24 & 58.06 & 7.72 & 172.71 \\
GAN & StyleGAN-XL~\citep{sauer2022stylegan} & & & 2.41 & 267.8 & 168M & 1 \\
Diff. + VAE$^*$ & DiT-XL/2~\citep{peebles2022scalable} & 12.03 & 105.3 & 3.04 & 240.8 & 675M  & 250  \\
Diffusion & ADM+Upsample~\citep{dhariwal2021diffusion}  & 9.96  & 121.8 & 3.85 & 221.7 & 731M & 2000 \\
% GAN & BigGAN-deep~\citep{brock2018large} & 8.43 & 177.9 & & & 160M  & 1 \\
Diffusion & RIN~\citep{jabri2023scalable} & 3.95 & 216.0 & & & 320M  & 1000 \\
Diffusion & simple diffusion~\citep{hoogeboom2023simple} & 3.54 & 205.3 & 3.02 & 248.7 & 2B & 512\\
Diffusion & VDM++~\citep{kingma2023vdm}  & 2.99 & 232.2 & 2.65 & 278.1  & 2B & 512 \\
\hdashline
% \multicolumn{8}{l}{\textit{Latent-Space Generative Models}} & Latent bpp \\

MLM + VQ & MaskGIT~\citep{chang2022maskgit} & 7.32 & 156.0 & & & 227M & 12  \\
% MLM + VQ & Token-Critic~\citep{lezama2022improved} & 6.80 & 182.1 &  &  & 368M & 36  \\
MLM + VQ & DPC+Upsample~\citep{lezama2023discrete}  & 3.62 & 249.4 & & & 619M & 72    \\
% MLM + VQ & DPC-light(5)+U~\citep{lezama2023discrete}  & 3.62 & 249.40 & & & 619M & 72 & 0.039   \\
\midrule
\multirow{2}{*}{MLM + \quantizername{}} & \multirow{2}{*}{\emph{\modelname{} (this paper)}} & 4.61 & 192.4 & & & \multirow{2}{*}{307M} & 12 \\
& & 3.07 & 213.1 & \textbf{1.91} & \textbf{324.3} &  & 64 \\
% & \emph{\modelname{} (this paper)} (40 bits) &  & & &  & 312M & & 0.039 \\
\bottomrule
\end{tabular}
}
\vspace{-6mm}
\end{table}

%% file: tables/compression.tex
\begin{wraptable}{r}{0.45\textwidth}
\vspace{-10mm}
    \centering
    \caption{\textbf{Video compression metrics}.
    % We compare the compression quality of our approach using three common quantitative metrics (PSNR, MS-SSIM, and LPIPS) at 0.0384 bpp, the bit rate of MAGVIT. The results show that our model outperforms MAGVIT on all metrics, and it outperforms all methods on LPIPS. Since PSNR and MS-SSIM correlate relatively poorly with subjective assessments of visual quality, it is not surprising that our method scores worse on these distortion metrics than HEVC and VVC.
    }
    \label{tab:compression}
    \vspace{-2mm}
    \resizebox{\linewidth}{!}{%
    \begin{tabular}{l@{\hspace{5pt}}c@{\hspace{5pt}}c@{\hspace{5pt}}c@{\hspace{5pt}}c}
    \toprule
    Method  & LPIPS$\downarrow$ & PSNR$\uparrow$ & MS-SSIM$\uparrow$ \\
    \midrule
    HEVC~\citep{sullivan2012overview} & 0.199& 30.10 & 0.943 \\ 
    VVC~\citep{vvc} & 0.153 & \textbf{32.65} & \textbf{0.966} \\
    \midrule
    MAGVIT~\citep{yu2022magvit} & 0.144 & 23.70 & 0.846 \\ 
    \emph{\modelname{} (this paper)} & \textbf{0.104} & 26.18 & 0.894 \\
    \bottomrule
    \end{tabular}
    }
    \vspace{-8mm}
\end{wraptable}

% magvit v1
% {'LPIPS': 0.14433361030154757,
%  "MS-SSIM/sRGB/R'G'B'": 0.8455671715630425,
%  "PSNR/sRGB/R'G'B'": 23.695795461697045}

% magvitv2-58219767
% {'LPIPS': 0.1441961204952664,
%  "MS-SSIM/sRGB/R'G'B'": 0.8187397972106935,
%  "PSNR/sRGB/R'G'B'": 23.121115535481767}

% magvitv2-59508633
% {'LPIPS': 0.08232382029427424,
%  "MS-SSIM/sRGB/R'G'B'": 0.9356586151970757,
%  "PSNR/sRGB/R'G'B'": 27.856519574652776}

%% file: tables/understanding.tex
\setlength\intextsep{20pt}
\begin{wraptable}{r}{0.5\textwidth}
% \vspace{-4mm}
\centering
\caption{\textbf{Video action recognition performance}\\(classification accuracy$\uparrow$ $\times$100).}
\label{tab:understanding}
\vspace{-2mm}
\resizebox{\linewidth}{!}{%
\begin{tabular}{@{}l@{\hspace{5pt}}c@{\hspace{10pt}}c@{\hspace{5pt}}c@{\hspace{5pt}}c@{}}
\toprule
\makecell[r]{Token as transformer's:} & Output & \multicolumn{3}{c}{Input}  \\
Tokenizer & SSv2 & SSv2 & K400 & K600 \\ \midrule
% VQVAE in \citet{wang2022bevt} & 67.10 & - & - & - \\
% 3D VQVAE \lu{in BEVT?} \lijun{it's ours} & 64.13 & 41.27 & 44.44 & 45.67 \\
3D VQ-VAE & 64.13 & 41.27 & 44.44 & 45.67 \\
% MAGVIT~\citep{yu2022magvit} & 67.22 & 57.34 / 82.80 & 72.29 / 89.56 & 74.65 / 91.22 \\
MAGVIT~\citep{yu2022magvit} & 67.22 & 57.34 & 72.29 & 74.65 \\
\emph{\modelname{} (this paper)} & \textbf{67.38} & \textbf{62.40} & \textbf{75.34} & \textbf{77.93} \\ \midrule
\textcolor{gray}{Raw pixel} & \textcolor{gray}{\cl{64.83}} & \textcolor{gray}{63.08} & \textcolor{gray}{76.13} & \textcolor{gray}{78.92} \\ 
\textcolor{gray}{\cl{HoG descriptor~\citep{wei2022masked}}} & \textcolor{gray}{\cl{65.86}} & \textcolor{gray}{\cl{n/a}} & \textcolor{gray}{\cl{n/a}} & \textcolor{gray}{\cl{n/a}} \\
\bottomrule
\end{tabular}
}
\vspace{-14mm}
\end{wraptable}
\vspace{-8mm}
% \begin{table}[tp]
%     \centering
%     \caption{\textbf{Video action recognition performance} (classification accuracy $\times 100$).}
%     \label{tab:understanding}
%     \begin{tabular}{lccccccc}
%     \toprule
%     & Codebook & $\gL_\mathit{GAN}$ & $\gL_\mathit{Perc.}$ & $\gL_\mathit{Comm.}$ & $\gL_\mathit{Entr.}$ & SSv2 Acc.$\uparrow$ & K400 Acc.$\uparrow$ \\ \midrule
%     \multicolumn{4}{l}{Target: \textbf{MAGVIT}~\citep{yu2022magvit}} \\
%     (VQGAN) & 1024 & \checkmark & \checkmark & \checkmark & \checkmark & \\ 
%     (VQVAE) & 1024 & & & \checkmark & \checkmark \\ 
%     \midrule
%     \multicolumn{4}{l}{Target: \textbf{\modelname{}} (this paper)} \\
%     (\quantizername GAN) & 1024 & \checkmark & \checkmark & \checkmark & \checkmark & \\ 
%     & 4096 & \checkmark & \checkmark & \checkmark & \checkmark & \\ 
%     & 16384 & \checkmark & \checkmark & \checkmark & \checkmark & \\ 
%     & 65536 & \checkmark & \checkmark & \checkmark & \checkmark & \\ \hdashline
%      & 16384 & & \checkmark & \checkmark & \checkmark & \\ 
%     (\quantizername VAE) & 16384 &  &  & \checkmark & \checkmark & \\ 
%     & 16384 & & &  & \checkmark & \\ 
%     (L2 only) & 16384 & & &  &  & \\ \bottomrule
%     \end{tabular}
% \end{table}
\leavevmode

%% file: tables/ablation.tex
\begin{table}[tp]
\vspace{-4mm}
\centering
\scriptsize
\caption{\textbf{Ablation study verifying key design choices}.}
\label{tab:ablation}
\vspace{-2mm}
\begin{subtable}{0.36\linewidth}
\centering
\caption{\label{tab:ablation_arch}Causal architectures on UCF-101. FID is calculated on the first frame.}

\begin{tabular}{@{}l@{\hspace{2pt}}c@{\hspace{2pt}}c@{\hspace{2pt}}c@{}}
\toprule
 & \#Params & FID$\downarrow$ & FVD$\downarrow$ \\ \midrule
\textcolor{gray}{MAGVIT} & \textcolor{gray}{39M} & \textcolor{gray}{n/a} & \textcolor{gray}{107.15}  \\
C-ViViT & 90M & 28.02 & 437.54 \\
C-ViViT + MAGVIT & 67M & 13.52  & 316.70 \\
\makecell[l]{\emph{\modelname{}}:\\\ \ Causal 3D CNN} & 58M & \textbf{7.06} & \textbf{96.33} \\ \bottomrule
\end{tabular}
\end{subtable}
\hfill
\begin{subtable}{0.3\linewidth}
\centering
\caption{\label{tab:ablation_imagenet}Image tokenization on\\ImageNet 128$\times$128.}
% \resizebox{\linewidth}{!}{%
\begin{tabular}{@{}l@{\hspace{2pt}}c@{\hspace{2pt}}c@{}}
\toprule
 & FID$\downarrow$ & LPIPS$\downarrow$ \\ \midrule
MAGVIT & 2.65 & 0.1292 \\
+ LFQ & 2.48 & 0.1182 \\
+ large vocabulary & 1.34 & 0.0821 \\
+ up/downsampler & 1.21 & 0.0790 \\
% + $\gL_{reconstruction}$ weight &  &  \\
+ deeper model & 1.20 & 0.0686 \\
+ adaptive normalization & \textbf{1.15} & \textbf{0.0685} \\ \bottomrule
\end{tabular}
% }
\end{subtable}
\hfill
\begin{subtable}{0.33\linewidth}
\centering
\caption{\label{tab:ablation_ucf101}Video tokenization on UCF-101.}
% \resizebox{\linewidth}{!}{%
\begin{tabular}{@{}l@{\hspace{2pt}}c@{\hspace{2pt}}c@{}}
\toprule
 & FVD$\downarrow$ & LPIPS$\downarrow$ \\ \midrule
MAGVIT & 24.55 & 0.0988 \\
+ LFQ \& large vocabulary & 16.12 & 0.0694 \\
+ up/downsampler & 15.37 & 0.0678 \\
+ late temporal downsample & 11.11 & 0.0653 \\
+ deeper model & 8.90 & 0.0542 \\
+ 3D blur pooling & \textbf{8.62} & \textbf{0.0537} \\ 
% + causal 3D CNN & \textbf{} & \textbf{} \\
\bottomrule 
\end{tabular}
% }
\end{subtable}
\vspace{-4mm}
\end{table}

%% file: sections/related.tex
\vspace{-4mm}
\paragraph{Visual tokenization.}
%Mapping visual inputs to compact discrete representations is a longstanding goal in machine learning.
Beyond the VQ-VAE models discussed in \cref{sec:background}, additional models have been proposed. ViT-VQGAN \citep{yu2021vector} introduces transformer blocks as a substitute for CNNs for image tokenization. C-ViViT \citep{villegas2022phenaki} further extends this idea for video tokenization. %Tokenizing video is more complex since it requires modeling the visual dynamics within the compressed spatial-temporal latent space. 
Early studies on video tokenization treat frames as independent images with no temporal compression~\citep{wu2021n,gupta2022maskvit}. Later research \citep{yan2021videogpt,ge2022long,yu2022magvit} integrates 3D CNNs to tokenize spatial-temporal volumes. Despite these advances in vector quantization (VQ), the codebook learned by previous VQ models is relatively small (\eg, $8$k) due to the difficulty in improving the generation quality with larger vocabularies. In contrast, our tokenizer can induce a large vocabulary (\eg, $262$k) that can be effectively modeled by an LM, leading to enhanced image and video generation quality.

% In the deep learning era, VQVAE \citep{van2017neural} shows the first learned discrete representations that outperform continuous ones for image generation, by proposing an extension of the VAE \citep{kingma2013auto} that uses a learned codebook for latent embedding quantization. VQGAN \citep{esser2021taming} extends the VQVAE with adversarial and perceptual losses to mitigate the blurriness induced by the reconstruction loss. While the original VQVAE used a PixelCNN \citep{van2016conditional} to model the prior of discrete latents, \cite{esser2021taming} use a language model for this task, achieving high resolution image generation.  

% While it is possible to tokenize each frame in a video separately as an image \citep{wu2021n},  TATS \citep{ge2022long} and MAGVIT \citep{yu2022magvit} incorporate 3D convolutional filters in the encoder and decoder of the VQGAN to tokenize videos as a whole. ViT-VQGAN \citep{yu2021vector} for images, and  C-ViViT \citep{villegas2022phenaki} for videos, show it is also possible to use a transformer backbone instead of a CNN for the VQGAN. 

% An intuitive solution is to simply decompose the codebook into several chunks along the channel dimension~\citep{rakhimov2021latent}, but this results in even longer sequences.

\vspace{-4mm}
\paragraph{Text-to-\{image, video\}.}
Text-to-image and text-to-video generation has garnered significant rapid advancements using both language models~\citep{yu2023scaling,chang2023muse} and diffusion models~\citep{ho2022imagen,blattmann2023align,singer2022make,ge2023preserve,ramesh2022hierarchical}. Although diffusion models, such as Midjourney, are considered the top performers in these tasks, it is unclear whether their advantage stems from the model, data, or some other unidentified factors. Indeed, it is challenging to scientifically compare these text-to-image models as they are trained on varied datasets, with some even being proprietary data, under inconsistent training conditions. To facilitate a fairer comparison, this paper prioritizes using the ImageNet and Kinetics benchmarks.
% We are training text-to-video models which may be ready by the next revision.

\vspace{-4mm}
\paragraph{Diffusion models.}
Exhibiting high quality sampling, pixel-space diffusion models \citep{sohl2015deep,song2019generative,ho2020denoising} raised to the top of the generative modeling space for both image \citep{ho2020denoising,dhariwal2021diffusion,saharia2022photorealistic} and video \citep{ho2022video,ho2022imagen,singer2022make} synthesis. The pixel-space denoising diffusion models (DDMs) are later refined by the latent-space DDM \citep{rombach2022high}, which conducts diffusion over the \emph{continuous} latent embeddings derived from a pre-trained variational autoencoder (VAE).
Binary latents for image modeling were used in \cite{wang2023binary}, where the diffusion process is parameterized with Bernoulli distributions.
%The stable diffusion model is a prominent example of latent DDM.
% Latent diffusion models for images \citep{rombach2022high} utilize the same modeling principle, but instead of pixel values model the distribution of.
%Conversely, while the U-Net~\citep{ronneberger2015u} has been the common neural architecture for both pixel- and latent-space diffusion models, 
Recent studies have identified advantages in substituting the U-Net~\citep{ronneberger2015u} denoising backbone with a Transformer \citep{peebles2022scalable,jabri2023scalable} or a hybrid of both \citep{hoogeboom2023simple}, making the distinctions between diffusion and language models in visual generation more blurred, with a key distinction being their latent format — continuous for diffusion and discrete for language models.
%Specifically, both DDM and MLM emulate a gradual corruption process on the latents and generate samples through iterative denoising restoration.
%In this paper we demonstrate that discrete latent space modeling outperforms all the compared continuous diffusion approaches, both in sampling quality and efficiency, when utilizing an appropriate visual tokenizer.

%% file: sections/conclusion.tex
We introduce \modelname{}, a novel video tokenizer that exploits lookup-free quantization along with architectural advancements to tokenize images and video with a shared vocabulary. The experiments show that our tokenizer outperforms the previously leading video tokenizer across three areas: visual generation, video compression, and action recognition in videos. Our results suggest that a good visual tokenizer is key for enabling language models to excel in image and video generation. These results demonstrate the great capabilities of LMs in visual generation, and advocate for further exploration of advanced visual tokenization methods designed for LLMs.
%We have demonstrated that Language Models outperform diffusion models for visual generation when equipped with a good visual tokenizer mapping pixel-space inputs to discrete tokens in a suitable vocabulary. To this end, we 
%\modelname{} outperforms previous video tokenizers in reconstruction and compression quality. When coupled with a Masked Language Model, it outperforms all previous methods, including diffusion models, on standard image and video generation benchmarks including ImageNet and Kinetics.

%% file: sections/appendix.tex
\section{Implementation Details}
\subsection{Image and Video Generation}

We set up two image tokenizers to downsample by 16$\times$ and 32$\times$, where they are used for generation at 256$\times$256 and 512$\times$512, respectively.
In both cases, an image is represented as 16$\times$16 tokens.
% to map an image into 16$\times$16 tokens.
% This represents a 32$\times$ downsample for the 512$\times$512 version and 16$\times$ for 256$\times$256.
% with 32$\times$ and 16$\times$ downsample ratio, respectively, for generation at 512$\times$512 and 256$\times$256 resolutions, where an image is always represented as 16$\times$16 tokens.
% The image tokenizer maps an image, either 256$\time$256 or 512$\times$512 resolution, into 16$\times$16 tokens.
We train them on the ImageNet training set for 270 epochs using a batch size of 256, both with 256$\times$256 images.

With this tokenizer we train a Masked Language Model following \citet{yu2022magvit}, using the token factorization described in \cref{sec:tokenizer_improvement}.
We train for 1080 epochs in accordance with the prior best model MDT~\citep{gao2023masked}, with batch size 1024 for better efficiency.
For preprocessing and data augmentation, we randomly crop 80-100\% of an image while keeping the aspect ratio, followed by random horizontal flipping.
% Data augmentation including random flipping 
% Following the literature on ImageNet, we do not apply data augmentation. 
% We resize the image keeping its aspect ratio and then proceed with a central crop.
The class label is dropped for 10\% of the training batches to enable classifier-free guidance~\citep{ho2021classifier}.
For unguided generation, we use temperature 30 for 512$\times$512 and 15 for 256$\times$256 in the non-autoregressive decoding.
For guided generation, we adopt the guidance schedule from \citet{gao2023masked} with temperature scaling~\citep{lezama2023discrete}, where we use guidance scale 25 with temperature 15.

We inflate an image tokenizer trained at 128$\times$128 for video modeling.
Different from the inflation in \citet{yu2022magvit}, we fill in the temporally last slice to correspond to the causal padding scheme. In addition, we disable the inflation for the discriminator and train it from scratch for better stability.
We train the causal video tokenizer on Kinetics-600 training set for 190 epochs with batch size 256.
This tokenizer is also used in subsequent evaluations of video compression and action recognition.

With the causal tokenizer producing 5$\times$16$\times$16 tokens for a 17$\times$128$\times$128 clip, the first 2$\times$16$\times$16 tokens are provided as the condition of the first 5 frames, per the standard setup of Kinetics-600 frame prediction benchmark.
We train the MLM transformer following \cite{yu2022magvit} with token factorization for 360 epochs with batch size 256.
The model is sampled with a cosine schedule using temperature 32.
% The videos 
% To use the inflation in~\citep{yu2022magvit} for the causal 3D layers, we fill in the temporally last slice to correspond to the causal padding scheme. In addition, we disable the inflation for the discriminator and train it from scratch for better stability.

% ImageNet is still considered one of the most crucial benchmarks because it allows for a consistent comparison of different models under consistent conditions, \eg, the same training data, similar training budget, and comparable model sizes. This differs from text-to-image tasks, where no standardized training data have been widely adopted for systematic comparisons.

% Our tokenizer can be used in text-to-vieo. However, due to the lack of time, we depriotize such comparison for the same . We hope to show some text-to-video results in the rebuttal.

\cl{
\subsection{Model Setup and Hyperparameters}
\label{app:hparams}
% \lijun{hparam tables}
\begin{figure}[tp]
    \centering
    \includegraphics[width=\linewidth]{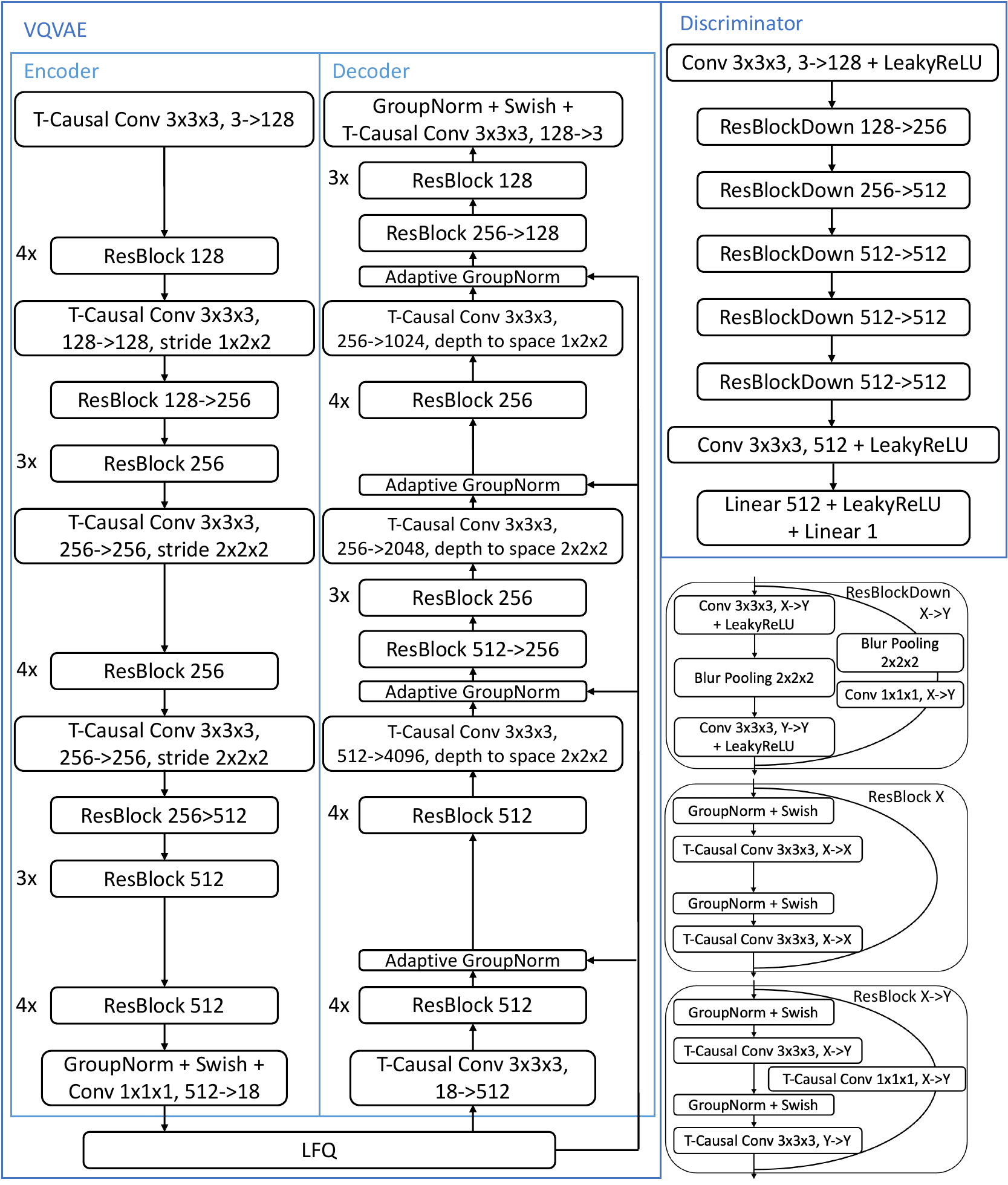}
    \caption{\cl{\textbf{\modelname{} tokenizer architecture}. T-Causal Conv refers to temporally causal convolution.}}
    \label{fig:full_arch}
\end{figure}
\cref{fig:full_arch} illustrates the architecture of our proposed \modelname{}.
We provide detailed training hyperparameters for our models as listed below:

% \begin{itemize}[nosep, leftmargin=*]
%     \item Video Tokenizer:
    \begin{itemize}[nosep, leftmargin=*]
    \item Video input: $17$ frames, frame stride $1$, $128\times128$ resolution.
    \item Base channels: $128$.
    \item VQVAE channel multipliers: $1, 2, 2, 4$. 
    \item Discriminator channel multipliers: $2, 4, 4, 4, 4$.
    \item Number of residual blocks: $4$.
    \item Latent shape: $5\times16\times16$.
    \item Vocabulary size:  $2^{18}$.
    \item Initialization: central inflation from a 2D model trained on ImageNet with this setup.
    \item Entropy loss weight: $0.1$.
    \item Entropy loss annealing steps: $2000$.
    \item Entropy loss annealing factor: $3$.
    \item Reconstruction loss weight: $5.0$.
    \item Generator loss type: Non-saturating.
    \item Generator adversarial loss weight: $0.1$.
    \item Discriminator gradient penalty: r1 with cost $10$.
    \item Perceptual loss weight: $0.1$.
    \item Commitment loss weight: $0.25$.
    \item LeCAM weight: $0.001$.
    \item Peak learning rate: $10^{-4}$.
    \item Learning rate schedule: linear warm up and cosine decay.
    \item Optimizer: Adam with $\beta_1=0$ and $\beta_2=0.99$.
    \item EMA model decay rate: $0.999$.
    \item Batch size: $256$.
    % \item Speed: 1.0 steps/sec on 256 TPU-v5e chips.
    \end{itemize}
    % \item Transformer:
    % \begin{itemize}[nosep, leftmargin=*]
    % \item Number of heads: $16$.
    % \item Number of layers: $24$.
    % \item Hidden size: $1024$.
    % \item MLP dimension: $4096$.
    % \item Vocabulary factorization groups: $2$.
    % \item Sequence length: $1281$.
    % \item Hidden dropout rate: $0.1$.
    % \item Attention dropout rate: $0.1$.
    % \item Mask rate schedule: cosine.
    % \item Peak learning rate: $10^{-4}$.
    % \item Learning rate schedule: linear warm up and cosine decay.
    % \item Optimizer: Adam with $\beta_1=0.9$ and $\beta_2=0.96$.
    % \item Weight decay $0.045$.
    % \item Label smoothing: $10^{-4}$.
    % \item Max gradient norm: $1$.
    % \item Batch size: $256$.
    % \item Speed: 3.7 steps/sec on 256 TPU-v5e chips.
% \end{itemize}
% \end{itemize}
}

\subsection{Video Compression Evaluation}

\begin{figure}[h]
    \includegraphics[width=\textwidth]{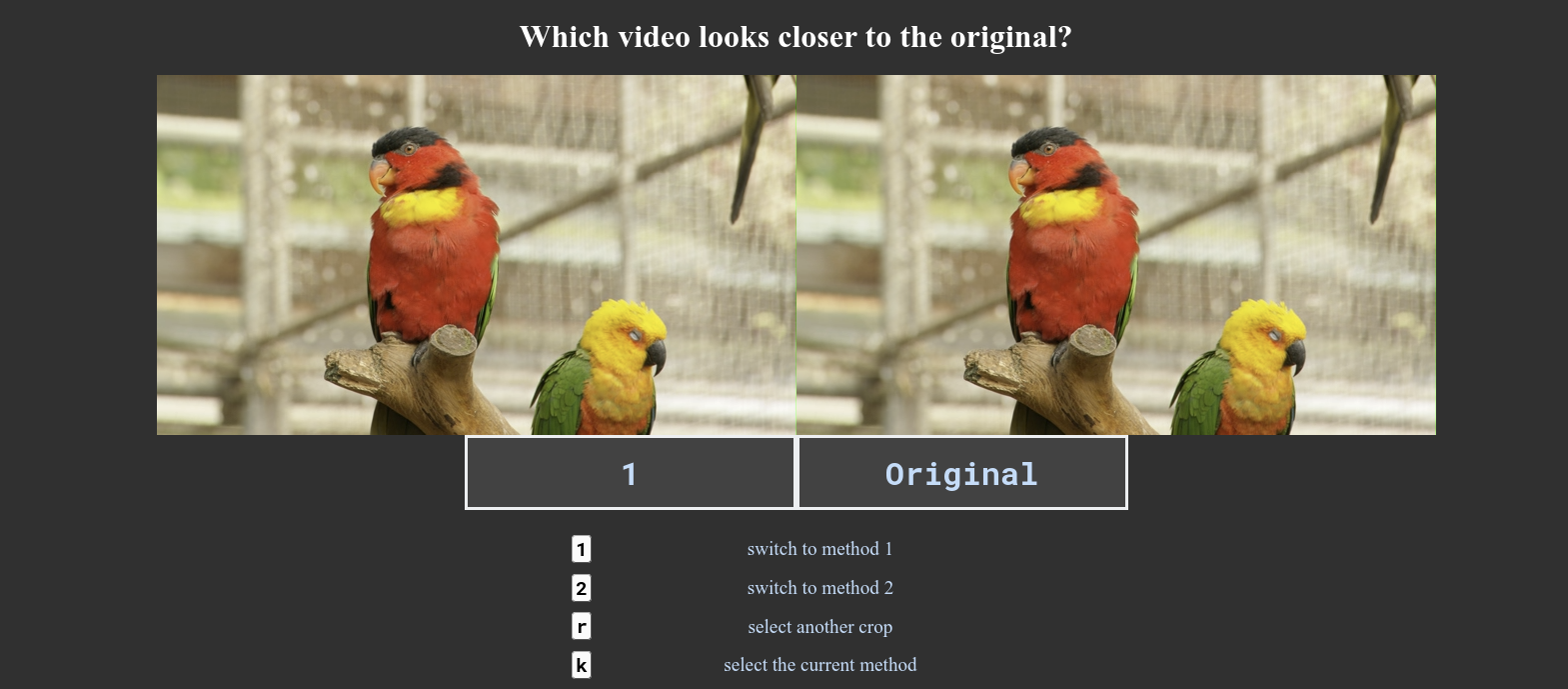}
    \caption{\textbf{Rating interface for subjective compression evaluation}.}
    \label{fig:user_rater}
\end{figure}

To rate the quality of the different methods, we use a two-alternative forced choice rating methodology~\citep{fechner1860elemente}. As this methodology produces a sequence of binary decisions, we calculate Elo scores~\citep{elo1978rating} based on pairwise preferences to quantify the relative visual quality between the models. The study was conducted on the 30 videos of the MCL-JCV dataset~\citep{wang2016mcl}, scaled down to a resolution of 640$\times$360 pixels.
Sixteen raters are engaged, each providing responses to an average of roughly 800 pairwise-preference questions. The questions are presented with an interface that parallels the one used for the Challenge on Learned Image Compression ({\small \url{http://compression.cc/}}), extended to comparing videos, as shown in \cref{fig:user_rater}. Raters are instructed to compare the two videos and are not allowed to pause the videos.

% \paragraph{Text-to-Image}

\subsection{Video Understanding Experiments}

\paragraph{Tokens as prediction targets.}
BEiT~\citep{bao2021beit} and BEVT~\citep{wang2022bevt} class of models pretrain visual encoders on pixel inputs by predicting tokens as targets in a masked-modeling framework, and demonstrate state-of-the-art downstream results. We use a simplified BEVT pre-training setup to test the effectiveness of our video tokens as targets for masked modeling. The main difference is that we drop the image-stream from pre-training and only use the video stream and for this reason, we also drop the multiple decoders completely and adopt an encoder-only architecture similar to BEiT. Detailed pre-training and fine-tuning setup is presented in \cref{tab:bevt_experiment_setup}. In \cref{tab:understanding} of the main paper, we show that our video tokens are effective targets for masked modeling based video understanding.

\paragraph{Tokens as inputs.}
In \cref{tab:understanding}, we show that we can re-use video understanding models trained on pixels using our video tokens as input, with very minimal performance drop. For this experiment, we train a factorized variant of the ViViT model \citep{arnab2021vivit} on pixels, and evaluate it on de-tokenized pixels from our model. We use the same hyper-parameters as used in \citet{arnab2021vivit} with a Base sized model operating on 32 frames of inputs at 224p resolution. For the Kinetics-600 experiment, we use the same hyper-parameters as the Kinetics-400 experiments.

\begin{table}[!t]
    \caption{\textbf{Experimental configurations with tokens as targets}.}
	\label{tab:bevt_experiment_setup}
	\centering
    %  \setlength\tabcolsep{4pt}
    % \resizebox{0.7\linewidth}{!}{
		\begin{tabular}	{l|l|l}
			\toprule
			\textbf{Config} & \textbf{SSv2 Pre-Training} & \textbf{SSv2 Fine-tuning} \\
			\midrule
    inputs & pixels & pixels \\
    input size & 16 $\times$ 224 $\times$ 224 $\times$ 3 & 16 $\times$ 224 $\times$ 224 $\times$ 3 \\
    targets & tokens & classes \\
    encoder & ViT-B & ViT-B \\
    decoder & linear & linear \\
    masking & block-tube \citep{wang2022bevt} & none \\
    masking ratio & 0.75 & 0.0 \\
    mask temporal length & 16 & 0 \\
	batch size & 1024  & 512  \\
	training epochs & 800 & 50  \\
	ViT sequence length & 8 $\times$ 16 $\times$ 16  & 8 $\times$ 16 $\times$ 16 \\
% 	optimizer
	\textbf{optimization} \\
	optimizer & AdamW  & AdamW \\ 
% 	base learning rate & 2.5e-1  \\
	optimizer momentum & 0.9 & 0.9 \\
	layer decay & 0.75  & 0.75 \\ 
	weight decay & 0.05 & 0.05 \\
	learning rate schedule & cosine decay  & cosine decay \\
	warmup epochs & 40  & 5 \\
	\textbf{data augmentations} \\
	random horizontal flip & true & false \\
	label smoothing & 0.1 & 0.1 \\
	mixup & none & 0.8 \\
	cutmix & none & 1.0 \\
	droppath & 0.0 & 0.1 \\
	dropout & 0.1 & 0.0 \\
	random color augmentation & false & false \\
		\bottomrule
		\end{tabular}
% 		}
\end{table}

\section{Additional Results}

\cl{
\begin{figure}
    \centering
    \begin{subfigure}{0.32\linewidth}
    \includegraphics[width=\linewidth]{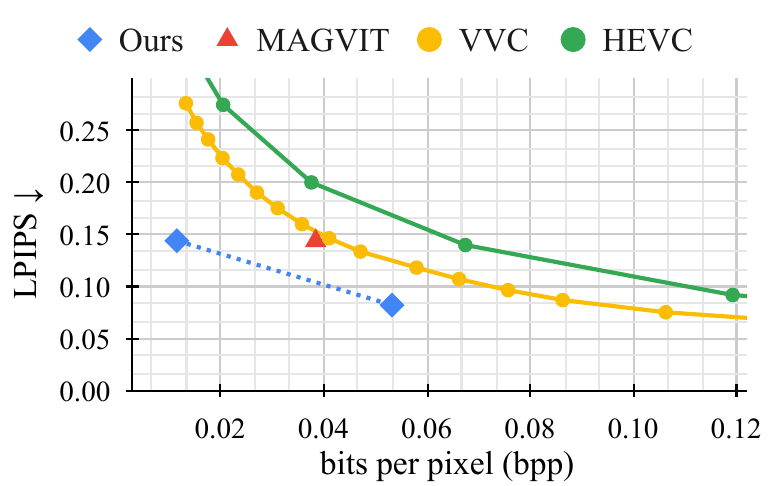}
    \caption{\cl{LPIPS$\downarrow$}}
    \end{subfigure}
    \begin{subfigure}{0.32\linewidth}
    \includegraphics[width=\linewidth]{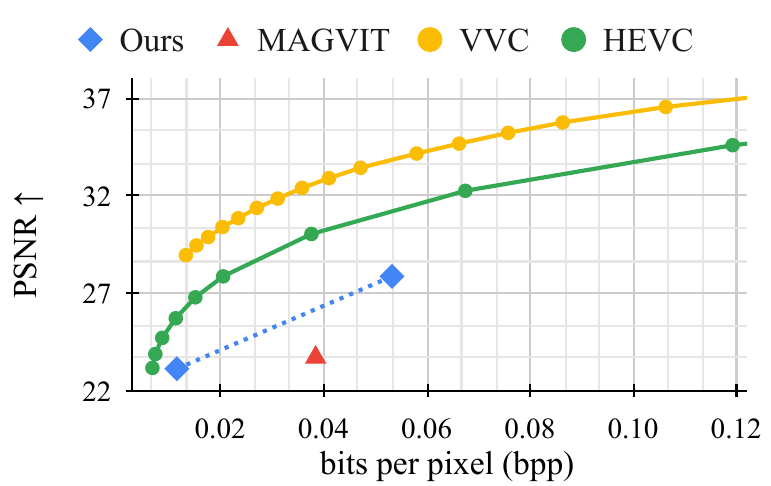}
    \caption{\cl{PSNR$\uparrow$}}
    \end{subfigure}
    \begin{subfigure}{0.32\linewidth}
    \includegraphics[width=\linewidth]{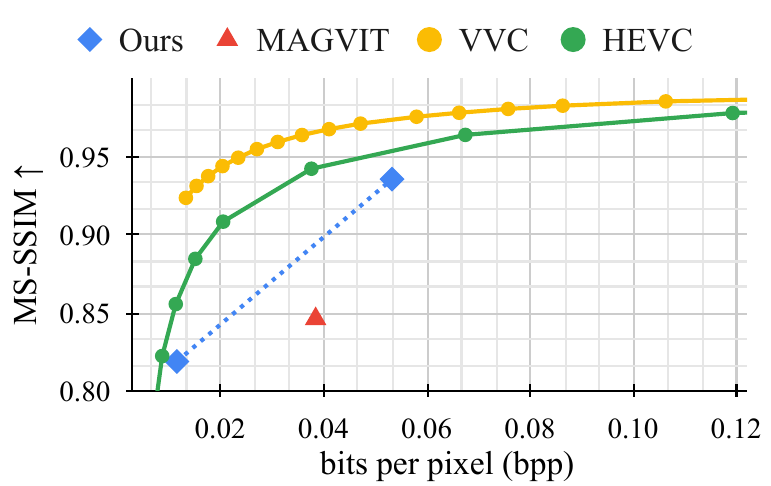}
    \caption{\cl{MS-SSIM$\uparrow$}}
    \end{subfigure}
    \caption{\cl{\textbf{Video compression metrics}, supplementary to \cref{tab:compression}.} }
    \label{fig:my_label}
\end{figure}
}

For better visualization, the generated video samples can be viewed at \webpage{}.

% \cl{
% \paragraph{Image generation on ImageNet.}
% \cref{tab:generation} shows the class-conditional image generation results on ImageNet 256$\times$256 benchmark, as discussed in \cref{sec:gen}.

% \paragraph{Autoregressive LM on UCF-101.}
% \cref{tab:gen_ar} shows the class-conditional video generation results on UCF-101 with AR-LM models, as discussed in \cref{sec:gen}.
% }

\paragraph{Where are the text-to-image results?} We want to emphasize that our goal is to develop a video tokenizer, and many of the proposed techniques are designed specifically for videos. Text-to-image may be out of the scope of our paper. We are currently training text-to-video models that require considerable computational resources. Due to time constraints, these results are not available at the moment. We intend to add the generated videos in the next revision. However, it is important to note that comparing these text-to-image or text-to-video models scientifically is challenging. These models were trained on different datasets, and some were even based on proprietary or non-public data, all under varying training conditions.

\input{tables/generation_bk}
\input{tables/generation_ar}

%% file: tables/generation_bk.tex
\begin{table}[tp]
\centering
\caption{\textbf{Class-conditional image generation on ImageNet 256$\times$256}. 
% \emph{Method types}: AutoRegressive Language Model (ARLM), Masked Language Model (MLM), Generative Adversarial Network (GAN), continuous Diffusion (Diff.). 
% \emph{Latent types}: \quantizerfullname{} (\quantizername{}), Vector Quantization (VQ), Variational AutoEncoder (VAE), Binary AutoEncoder (BAE).
% We report the bits per pixel (bpp) of the representation used in each latent-space models.
Guidance indicates the classifier-free diffusion guidance~\citep{ho2021classifier}.
$^*$ indicates usage of extra training data.
We adopt the evaluation protocol and implementation of ADM.
}
\label{tab:generation}

\resizebox{\linewidth}{!}{%
\begin{tabular}{@{}l@{\hspace{5pt}}l@{\hspace{4pt}}c@{\hspace{2pt}}c@{\hspace{3pt}}c@{\hspace{2pt}}c@{\hspace{2pt}}c@{\hspace{2pt}}c@{}}
\toprule
\multirow{2}{*}{Type} & \multirow{2}{*}{Method}  & \multicolumn{2}{c}{w/o guidance} & \multicolumn{2}{c}{w/ guidance}  & \multirow{2}{*}{\# Params}  & \multirow{2}{*}{Steps} \\
& & FID$\downarrow$ & IS$\uparrow$ & FID$\downarrow$ & IS$\uparrow$ \\ \midrule
% \multicolumn{6}{l}{\textbf{512 $\times$ 512 resolution}} \\
% % ADM & & 23.24 & 58.06 & 7.72 & 172.71 \\
% $[C]$ ADM-U  & 9.96  & 121.78 & 3.85 & 221.72 & 731M & 128$^2$ pixel & 2000 \\
% $[G]$ BigGAN-deep & 8.43 & 177.90 & & & 160M & & 24 & 1 \\
% $[C]$ RIN & 3.95 & 216.00 & & & 320M & & 24 & 1000 \\
% $[C]$ Simple diffusion & 3.54 & 205.30 & 3.02 & 248.70 & & & 24 &\\
% $[C]$ VDM++ & 2.99 & 232.20 & 2.65 & 278.10  & & & 24 &\\
% $[G]$ StyleGAN-XL & & & 2.41 & 267.75 & 168M & & 1 \\
% \hdashline
% $[C]$ DiT-XL/2 & 12.03 & 105.25 & 3.04 & 240.82 & 675M & 64$^2\times$(VAE 4)$^*$ & 250 \\
% $[T]$ MaskGIT  & 7.32 & 156.00 & & & 227M & 32$^2\times$(VQ 1024) & 12 \\
% % Token-Critic & 32$\times$32$\times$(VQ 1024) & 6.80 & 182.10  \\
% $[D]$ DPC  & 6.09 & 228.10 & & & 454M & 32$^2\times$(VQ 1024) & 180  \\
% \hdashline
% $[T]$ \emph{\modelname} (Ours)  & 3.51 & 235.47 & \textbf{1.92} & \textbf{325.02} & 307M & 16$^2\times$(\quantizername{} $2^{18}$) & 64 \\
% % $[T]$ \emph{\modelname}-XL (Ours) & & & & & 635M & 16$^2\times$(\quantizername{} $2^{18}$) \\
% \midrule
% \multicolumn{6}{l}{\textbf{256 $\times$ 256 resolution}} \\
% ADM &  & 10.94 & 100.98 & 4.59 & 186.70 \\
% \multicolumn{5}{l}{\textit{Pixel-Space Generative Models}} \\
GAN & BigGAN-deep~\citep{brock2018large} & 6.95 & 171.4 & & & 160M & 1 \\
GAN & StyleGAN-XL~\citep{sauer2022stylegan} & & & 2.30& 265.1 & 166M & 1 \\ 
\hdashline
Diff. + VAE$^*$ & LDM-4~\citep{rombach2022high} & 10.56  & 103.5 & 3.60 & 247.7 & 400M & 250 \\
Diff. + VAE$^*$ & DiT-XL/2~\citep{peebles2022scalable}  & 9.62  & 121.5 & 2.27  & 278.2 & 675M  & 250 \\
Diff. + BAE & Binary latent diffusion~\citep{wang2023binary} & 8.21 & 162.3 & & & 172M & 64  \\
Diffusion & ADM+Upsample~\citep{dhariwal2021diffusion} & 7.49 & 127.5 & 3.94  & 215.8 & 608M & 2000 \\
Diff. + VAE$^*$ & MDT~\citep{gao2023masked}  & 6.23  & 143.0 & 1.79  & 283.0 & 676M & 250  \\
Diff. + VAE$^*$ & MaskDiT~\citep{zheng2023fast} & 5.69  & 178.0 & 2.28  & 276.6 & 736M & 40 \\ 
Diffusion & CDM~\citep{ho2022cascaded}  & 4.88 & 158.7 & & & & 8100 \\
Diffusion & RIN~\citep{jabri2023scalable} & 3.42 & 182.0 & & & 410M &  1000 \\
Diffusion & simple diffusion~\citep{hoogeboom2023simple} & 2.77 & 211.8 & 2.44 & 256.3 & 2B & 512 \\
Diffusion & VDM++~\citep{kingma2023vdm} & 2.40 & 225.3 & 2.12 & 267.7 & 2B & 512 \\
\hdashline
% \multicolumn{8}{l}{\textit{Latent-Space Generative Models}} & Latent bpp \\
AR-LM + VQ & VQGAN~\citep{esser2021taming} & 15.78 & 78.3  & & & 1.4B & 256 \\
MLM + VQ & MaskGIT~\citep{chang2022maskgit}  & 6.18  & 182.1 & & & 227M & 8  \\
MLM + VQ & Token-Critic~\citep{lezama2022improved} & 4.69 & 174.5 & & & 368M & 36 \\
MLM + VQ & Contextual RQ-Transformer~\citep{lee2022draft} & 3.41 & 224.6 & & & 1.4B & 72 \\
MLM + VQ & DPC~\citep{lezama2023discrete}  & 4.45 & 244.8 & & & 454M & 180 \\
% ADM & - & 10.94 & 6.02 & 100.98 & 0.69 & 0.63 \\
\midrule
MLM + \quantizername{} & \emph{\modelname{} (this paper)} & 3.65 & 200.5 & \textbf{1.78} & \textbf{319.4} & 307M & 64  \\
% $[T]$ \emph{\modelname}-XL (Ours)  & $<$3 & & $<$2 & & 635M & 16$^2\times$(\quantizername{} $2^{18}$) \\
\bottomrule
\end{tabular}
}
% \end{subtable}
\end{table}\textbf{}

%% file: tables/generation_ar.tex
\begin{table}[tp]
\centering
% \vspace{-6mm}
\caption{\cl{\textbf{Video generation results}: class-conditional generation on UCF-101 with AR-LM models. We use the same transformer configuration as MLM experiments but without vocabulary factorization and weight tying. As a result, the AR-LM with \modelname{} uses more parameters in the embedding table and the softmax layer.}
% \emph{Method types}: AutoRegressive Language Model (ARLM), Masked Language Model (MLM), Generative Adversarial Network (GAN), continuous Diffusion (Diff.). 
% \emph{Latent types}: \quantizerfullname{} (\quantizername{}), Vector Quantization (VQ), Variational AutoEncoder (VAE), Discrete Cosine Transformation (DCT). 
% We report the bits per pixel (bpp) of the representation used in each latent-space model.
}
\label{tab:gen_ar}
% \vspace{-2mm}
% \resizebox{\linewidth}{!}{%
\centering
\cl{\begin{tabular}{@{}llcc@{}}
\toprule
Tokenizer   & FVD$\downarrow$  & \#Params  & \#Steps  \\ \midrule
MAGVIT~\citep{yu2022magvit} & 265  & 306M & 1024 \\
% &      MLM  & 76 & 306M   & 12  \\ \hdashline
\emph{\modelname{} (this paper)} & \textbf{109}  & 840M & 1280\\
%  & MLM &  \textbf{58}  & 307M & 12  \\
\bottomrule
\end{tabular}
}
% \vspace{-2mm}
% }
\end{table}